\pdfoutput=1
\documentclass[11pt]{article}
\usepackage[]{coling}
\usepackage{times}
\usepackage{latexsym}
\usepackage{float}
\usepackage[T1]{fontenc}
\usepackage[utf8]{inputenc}
\usepackage{microtype}
\usepackage{inconsolata}
\usepackage{graphicx}
\usepackage{xcolor}
\usepackage{soul}
\usepackage{colortbl}
\usepackage{color}
\usepackage{enumitem}
\usepackage{fontawesome5}
\usepackage{tabularx}
\usepackage{dingbat}
\usepackage{pifont}
\usepackage{amsmath}
\usepackage{amssymb}
\usepackage{makecell}
\usepackage{arydshln}
\usepackage{stfloats}
\usepackage{subfig}
\usepackage{CJKutf8}
\usepackage{multirow}
\usepackage{array}
\usepackage[utf8]{inputenc}
\usepackage{microtype}
\usepackage{booktabs}
\usepackage{xcolor}
\usepackage{fontawesome5}
\usepackage{caption}
\usepackage{amsfonts}
\usepackage{inconsolata}
\usepackage{arydshln}
\usepackage{fancyvrb}
\usepackage{listings}
\usepackage{grffile}
\usepackage{pifont}
\usepackage{tikz}

\newcommand{\cmark}{{\textbf{\textcolor[rgb]{0.1, 0.5, 0.1}{\ding{51}}}}}
\newcommand{\xmark}{{\textbf{\color{red}{\ding{55}}}}}
\definecolor{myblue}{rgb}{0.82, 0.94, 0.75}
\definecolor{mygold}{rgb}{1, 0.92, 0.56}
\definecolor{mylightblue}{rgb}{0.70, 0.83, 0.96}
\definecolor{mylightyellow}{rgb}{0.96, 0.88, 0.49}
\definecolor{mylightpink}{rgb}{0.93, 0.79, 0.80}

\newcommand{\hbarthree}[3]{
    \begin{tikzpicture}[scale=0.04, inner sep=0pt, outer sep=0pt]
        \fill[mylightblue] (0,0) rectangle (#1,8);  
        \fill[mylightyellow] (#1,0) rectangle ({#1+#2},8);  
        \fill[mylightpink] ({#1+#2},0) rectangle ({#1+#2+#3},8);  
        \node at (#1/2, 4) {\scriptsize #1\%};
        \node at ({#1+#2/2}, 4) {\scriptsize #2\%};
        \node at ({#1+#2+#3/2}, 4) {\scriptsize #3\%};
    \end{tikzpicture}
}

\newcommand*\colourcheck[1]{%
  \expandafter\newcommand\csname #1check\endcsname{\textcolor{#1}{\ding{52}}}%
}
\colourcheck{blue}
\colourcheck{green}
\colourcheck{red}
\colourcheck{black}
\usepackage{CJKutf8}

\usepackage{highlight}
\newcolumntype{a}{>{\columncolor{Gray}}c}

\usepackage[nointegrals]{wasysym}
\definecolor{darkseagreen}{rgb}{0.46, 0.74, 0.46}
\definecolor{alizarin}{rgb}{0.82, 0.1, 0.26}
\usepackage{CJKutf8}

\usepackage{amssymb}
\usepackage{pifont}

\definecolor{darkblue}{rgb}{0,0,.5}
\definecolor{darkgreen}{rgb}{0,.5,0}
\definecolor{lightgray}{rgb}{.8,.8,.8}
\definecolor{aliceblue}{rgb}{0.75, 0.75, 1.0}
\definecolor{darkseagreen}{rgb}{0.46, 0.74, 0.46}
\definecolor{alizarin}{rgb}{0.82, 0.1, 0.26}
\definecolor{airforceblue}{rgb}{0.36, 0.54, 0.66}
\definecolor{red_graph}{rgb}{0.98, 0.8, 0.8}
\definecolor{blue_graph}{rgb}{0.8, 0.98, 0.8}
\definecolor{red}{rgb}{0.8, 0.0, 0.0}
\definecolor{burgundy}{rgb}{0.5, 0.0, 0.13}
\definecolor{britishracinggreen}{rgb}{0.0, 0.26, 0.15}

\definecolor{CustomBlue}{RGB}{57,83,191}
\usepackage[most]{tcolorbox}
\newtcbox{\clustertab}[1]{on line, box align=base, colback={#1},colframe={#1},size=fbox,arc=2pt,top=-1.5pt, bottom=-1.5pt, left=-1.5pt, right=-1.5pt, boxrule=0pt, enlarge left by=1pt}

\definecolor{Gray}{gray}{0.85}

\usepackage{colortbl}

\newcolumntype{a}{>{\columncolor{Gray}}c}
\definecolor{set10-blue}{HTML}{4169E1}
\definecolor{google-red}{HTML}{de5246}

\newcommand{\win}[1]{\cellcolor{blue!10}{#1}{ \hspace{0.1cm}\small\clustertab{blue!70}{\color{blue!1} $\mathbf{1}$}}}
\newcommand{\two}[1]{\cellcolor{darkseagreen!15}{#1}{ \hspace{0.1cm}\small\clustertab{darkseagreen!100}{\color{darkseagreen!1} $\mathbf{2}$}}}
\newcommand{\thi}[1]{\cellcolor{orange!15}{#1}{ \hspace{0.1cm}\small\clustertab{orange!90}{\color{orange!1} $\mathbf{3}$}}}
\newcommand{\fou}[1]{\cellcolor{red!15}{#1}{ \hspace{0.1cm}\small\clustertab{red!80}{\color{red!1} $\mathbf{4}$}}}
\newcommand{\six}[1]{\cellcolor{gray!15}{#1}{ \hspace{0.1cm}\small\clustertab{gray!90}{\color{gray!1} $\mathbf{6}$}}}
\newcommand{\fiv}[1]{\cellcolor{olive!15}{#1}{ \hspace{0.1cm}\small\clustertab{olive!90}{\color{olive!1} $\mathbf{5}$}}}

\title{MQM-APE: Toward High-Quality Error Annotation Predictors with Automatic Post-Editing in LLM Translation Evaluators}

\author{Qingyu~Lu$^{\heartsuit}$,
\ Liang Ding$^{\Re}$, 
\ Kanjian Zhang$^{\heartsuit\spadesuit}$\thanks{~~Corresponding Author.},
\ Jinxia Zhang$^{\heartsuit}$,
\ Dacheng Tao$^{\diamondsuit}$ \\
\small \ $^{\heartsuit}$Southeast University
\ $^{\Re}$The University of Sydney
\ $^{\spadesuit}$Southeast University Shenzhen Research Institute \\
\small \ $^{\diamondsuit}$College of Computing and Data Science at Nanyang Technological University, Singapore 639798
\\
\includegraphics[scale=0.14]{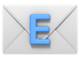}\texttt{ \small \{luqingyu,jinxiazhang,kjzhang\}@seu.edu.cn,} \texttt{\small \{liangding.liam,dacheng.tao\}@gmail.com}
 \\
\includegraphics[scale=0.025]{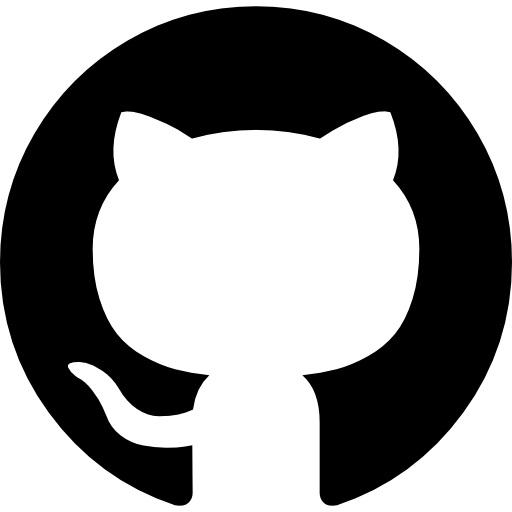} \ 
\small \url{https://github.com/Coldmist-Lu/MQM_APE}
}

\begin{document}
\maketitle
\begin{abstract}

Large Language Models (LLMs) have shown significant potential as judges for Machine Translation (MT) quality assessment, providing both scores and fine-grained feedback. Although approaches such as GEMBA-MQM \citep{kocmi-federmann-2023-gemba} have shown state-of-the-art performance on reference-free evaluation, the predicted errors do not align well with those annotated by human, limiting their interpretability as feedback signals. To enhance the quality of error annotations predicted by LLM evaluators, we introduce a universal and training-free framework, \textbf{MQM-APE}, based on the idea of filtering out non-impactful errors by Automatically Post-Editing (APE) the original translation based on each error, leaving only those errors that contribute to quality improvement. Specifically, we prompt the LLM to act as \ding{182} \textit{evaluator} to provide error annotations, \ding{183} \textit{post-editor} to determine whether errors impact quality improvement and \ding{184} \textit{pairwise quality verifier} as the error filter. Experiments show that our approach consistently improves both the reliability and quality of error spans against GEMBA-MQM, across eight LLMs in both high- and low-resource languages. Orthogonal to trained approaches, MQM-APE complements translation-specific evaluators such as Tower, highlighting its broad applicability. Further analysis confirms the effectiveness of each module and offers valuable insights into evaluator design and LLMs selection.

\end{abstract}

\section{Introduction} 

Machine Translation (MT) evaluators assess translation quality and play a key role in aligning with human judgements \citep{freitag-etal-2022-results, lu-etal-2023-toward}, especially in the era of Large Language Models (LLMs, \citealp{achiam2023gpt, touvron2023llama, peng-etal-2023-towards}). As the demand for interpretability grows \citep{xu-etal-2023-instructscore, leiter-etal-2023-eval4nlp, leiter2024towards}, these evaluators offer fine-grained, reflective feedback to enhance translation quality in scenarios such as automatic post-editing (APE, \citealp{ki-carpuat-2024-guiding}) or preference alignment training \citep{ramos2023aligning, he-etal-2024-improving}.

\begin{table}[t]
 \centering
\renewcommand\arraystretch{1.2}
 \setlength{\tabcolsep}{3pt}{
 \resizebox{\linewidth}{!}{\begin{tabular}{ccccc}\toprule
   \makecell{\large\textbf{Error-based MT Evaluation}} & \makecell{\textbf{Fine-grained}\\\textbf{Feedback}} & \makecell{\textbf{Error Span}\\\textbf{Enhancement}} &  \makecell{\textbf{Post-Edited}\\\textbf{Translation}} \\\midrule
   \multicolumn{4}{c}{\textit{Training-Dependent (Resource-Limited) Approaches }} \\\hdashline
   InstructScore \citep{xu-etal-2023-instructscore} & \cmark & \cmark & \xmark \\
   xCOMET \citep{guerreiro2023xcomet} & \cmark & \cmark & \xmark \\
   LLMRefine \citep{xu-etal-2024-llmrefine} & \cmark & \xmark & \cmark \\
   Tower \citep{alves2024tower} & \cmark & \xmark & \cmark \\\midrule
   \multicolumn{4}{c}{\textit{Training-Free (Model-Agnostic) Approaches}} \\\hdashline
   GEMBA \citep{kocmi-federmann-2023-large} & \xmark & \xmark & \xmark \\
   EAPrompt \citep{lu-etal-2024-error} & \cmark & \xmark & \xmark \\
   AutoMQM \citep{fernandes-etal-2023-devil} & \cmark & \xmark & \xmark \\
   GEMBA-MQM \citep{kocmi-federmann-2023-gemba} & \cmark & \xmark & \xmark
   \\\midrule\rowcolor{gray!16}
   \textbf{\large{MQM-APE (ours)}} & \cmark & \cmark & \cmark   
   \\\bottomrule
 \end{tabular}}}
 \caption{\textbf{Related work on error-based MT evaluation}. MQM-APE is a training-free approach that improves upon GEMBA-MQM \citep{kocmi-federmann-2023-gemba} and complements training-dependent approaches such as Tower \citep{alves2024tower}. It offers high-quality error annotations and post-edited translations.}
 \vspace{-10pt}
 \label{tab:relatedworks}
\end{table}

The success of LLMs in reasoning and generation highlights their potential for MT quality assessment \cite{kocmi-federmann-2023-large}. Prompt strategies take advantage of Multidimensional Quality Metrics (MQM, \citealp{Lommel2018,freitag-etal-2021-experts}), an error-based human evaluation framework, to provide both scores and explicit error annotations \citep{lu-etal-2024-error}. Although these evaluators, such as GEMBA-MQM, have shown SOTA performance in reference-free evaluation \citep{kocmi-federmann-2023-gemba}, the predicted error spans are not well aligned with human judgements \citep{fernandes-etal-2023-devil, huang-etal-2024-lost}. Training-dependent approaches \citep{xu-etal-2023-instructscore, guerreiro2023xcomet} face high computational costs, limiting their applicability across models and languages, as shown in Table~\ref{tab:relatedworks}.

Building upon training-free approaches, we propose a universal framework, \textbf{MQM-APE}, that significantly enhances evaluator performance and improves the quality of predicted error spans. Inspired by findings that APE or self-refinement can enhance translation quality through fine-grained feedback with error spans \citep{ki-carpuat-2024-guiding, xu-etal-2024-llmrefine}, we integrate APE into the translation evaluation process to refine the set of errors. As shown in Figure~\ref{fig:overview}, the idea is to \textit{filter out non-impactful errors by post-editing the original translation based on each error, leaving only those errors that contribute to quality improvement.}

Specifically, we prompt the LLM to act as \ding{182} \textit{evaluator} to provide error annotations (\S\ref{sec:approach_stage1}), \ding{183} \textit{post-editor} to determine whether errors impact quality improvement (\S\ref{sec:method_ape}) and \ding{184} \textit{pairwise quality verifier} as the error filter (\S\ref{sec:method_verifier}). This evaluation pipeline preserves only \textbf{genuinely impactful errors} that contribute to quality improvements after APE and influence the final score, as translation quality is assessed based on error count (\S\ref{sec:method_postprocess}).

Extensive experiments are conducted on eight different instruction-tuned LLMs using the WMT22 test set \citep{freitag-etal-2022-results}, which includes 106,758 segments from 54 MT systems, as well as the IndicMT test set with 1,000 MQM-annotated segments in Indian languages \cite{sai-b-etal-2023-indicmt}. Our findings reveal that:

\begin{figure*}[t]
\centering
\includegraphics[scale=0.4]{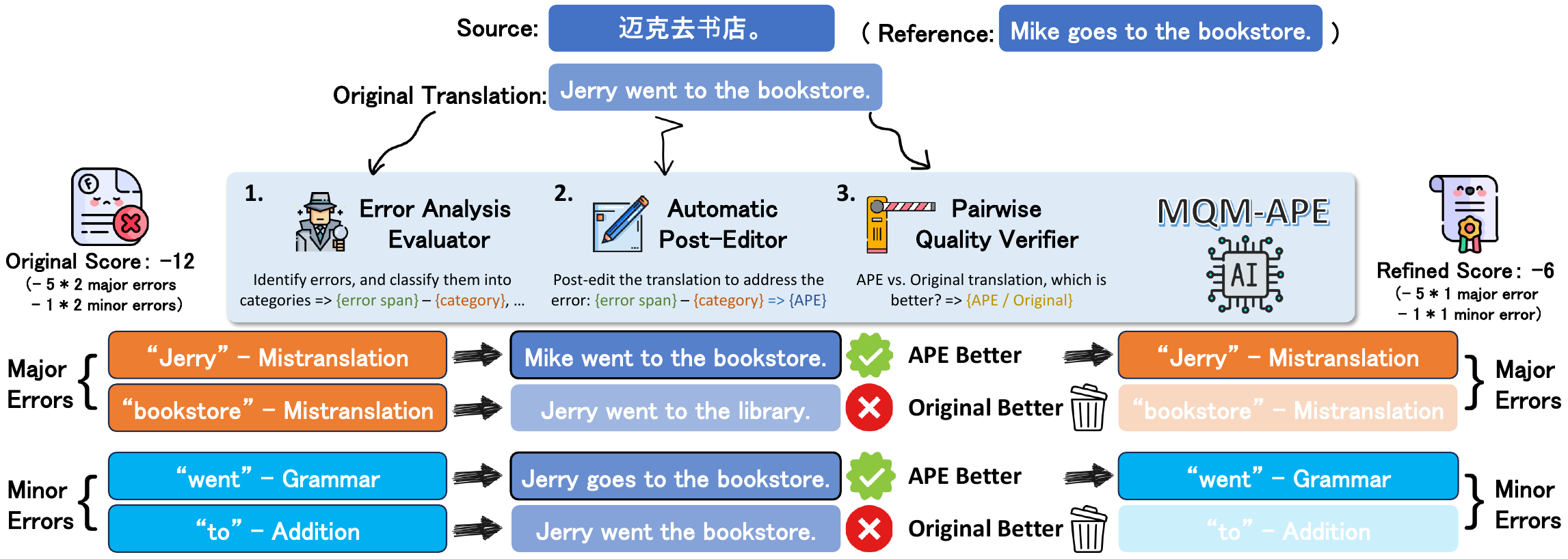}
\caption{\textbf{A comparative overview of our MQM-APE approach}. The evaluated translation passes through three sequential modules, all operated by the same LLM: 1) the Error Analysis Evaluator, which provides detailed error demonstrations; 2) APE, which post-edits the translation based on each error annotation; and 3) the Pairwise Quality Verifier, which verifies whether quality improves after post-editing.}
\label{fig:overview}
\end{figure*}

\begin{itemize}[left=0mm,itemsep=0mm,topsep=0.2mm]
  \item MQM-APE surpasses GEMBA-MQM \citep{kocmi-federmann-2023-gemba} at both the system and segment levels, offering interpretable error spans that closely align with human annotations.
  \item MQM-APE generalizes across a broad range of LLMs and is effectively applicable to both high- and low-resource languages.
  \item Orthogonal to training-dependent approaches, MQM-APE complements translation-specific evaluators such as Tower \citep{alves2024tower}.
  \item APE translations exhibit superior overall quality compared to the original translations (\S\ref{sec:res_postedit}).
  \item Quality Verifier aligns with modern metrics like $\text{CometKiwi}_{22}^{\text{QE}}$ (\S\ref{sec:res_verifier}), which can be replaced by these metrics with comparable effects (\S\ref{sec:replace_metrics}).
  \item MQM-APE introduces acceptable costs against GEMBA-MQM (\S\ref{sec:inference_cost}), and preserves error distribution across severities and categories (\S\ref{sec:error_dist}).
\end{itemize}
Finally, we present a performance ranking of various backbone LLMs as translation evaluators (\S\ref{sec:recommend}), providing guidance for researchers navigating the trade-offs between reliability, interpretability, and inference cost.

\section{Preliminaries}

\paragraph{Translation Evaluation} Translation evaluation metrics assess MT quality using test sets, typically relying on the source, translation, and human references. This paper focuses on the reference-free scenario, where no human references are provided. The output is a score reflecting translation quality.

\paragraph{Human Evaluation} Human annotation serves as the gold standard for translation evaluation. Recently, the Multi-dimensional Quality Metric (MQM, \citealp{Lommel2018}) has been adopted by WMT\footnote{\url{https://www2.statmt.org/}} as a high-quality human evaluation framework \citep{freitag-etal-2022-results}. MQM requires categorising translation errors into "Critical", "Major" and "Minor" based on severity. The final score is calculated by weighting the number of errors according to their severity. 

{\ding{43}} See Appendix~\ref{appendix:MQM} for an introduction of MQM.

\paragraph{Error Analysis} Error analysis simulates human evaluations by predicting error demonstrations for interpretable assessments \citep{lu-etal-2023-toward}. Previous studies explore prompting strategies to enable LLMs to generate explicit error demonstrations \citep{fernandes-etal-2023-devil}. For instance, GEMBA-MQM \citep{kocmi-federmann-2023-gemba}, a state-of-the-art error analysis evaluator that employs 3-shot prompting strategies, serves as our baseline and main module in MQM-APE (\S\ref{sec:approach_stage1}).

\section{Methodology}

As shown in Figure~\ref{fig:overview}, we employ MQM-APE by prompting the same LLM to perform multiple roles without fine-tuning for each task. MQM-APE evaluates a given translation $y$ of source $x$ through three sequential modules: \ding{182} \textbf{Error Analysis Evaluator} identifies errors in $y$, providing error demonstrations $\mathcal{E}$ with error and severity; \ding{183} \textbf{Automatic Post Editor} post-edits $y$ based on each identified error $e_i \in \mathcal{E}$, producing a set of corrected translations $\mathcal{Y}_{\text{pe}}$; \ding{184} \textbf{Pairwise Quality Verifier} checks whether the post-edited translations improve upon the original translation $y$. Errors for which the APE translation fails to improve on the original are discarded, leaving a refined set of errors $\mathcal{E}^* \subseteq \mathcal{E}$ that contribute to quality improvement. The translation is finally scored based on the refined set of errors $\mathcal{E}^*$.

\subsection{Module 1: Error Analysis Evaluator} \label{sec:approach_stage1}

We follow \citet{kocmi-federmann-2023-gemba, lu-etal-2024-error} to prompt the LLM to perform MQM-like assessment, identifying errors in translation $y$ of source $x$. This step can be described as:
\begin{equation}
    \mathcal{E} = \text{Evaluator}(x, y),
\end{equation}
where $\mathcal{E} = \{e_1, e_2, \cdots, e_N\}$, represents the set of errors identified by the evaluator, and $N$ denotes the number of errors. 

For each error annotation, three types of information are recorded: the error span, which specifies the location of the error in $y$; the error category, such as mistranslation, omission, or grammatical issues, aligned with MQM guidelines \citep{freitag-etal-2021-experts}; and the error severity, classified as "Critical", "Major" or "Minor", reflecting their impact from highest to lowest. Note that translations without errors identified will bypass the subsequent steps.

\subsection{Module 2: Automatic Post-Editor} \label{sec:method_ape}

The purpose of APE is to correct errors in the original translation $y$ based on the given source segment $x$. We prompt the LLM to post-edit the translation for each identified error, using the error span and category as feedback, for instance, the span "Jerry" with the category "Mistranslation", as shown in Figure 1, can guide the LLM to edit the translation from "Jerry went to the bookstore" to "Mike went to the bookstore". This step can be formalized as:
\begin{equation}
    y_i^{(\text{pe})} = \text{APE}(x, y, e_i), \ \  i = 1, 2, \cdots, N
\end{equation}
where a set of post-edited translations $\mathcal{Y}_{\text{pe}}=\{y_1^{(\text{pe})},y_2^{(\text{pe})}, \cdots, y_N^{(\text{pe})}\}$ is produced.

\subsection{Module 3: Pairwise Quality Verifier} \label{sec:method_verifier}
To assess the impact of errors with APE, we prompt the LLM to act as a verifier, comparing the quality of each post-edited translation $y_i^{(\text{pe})}$ with original $y$. This step is expressed as:
\begin{equation}
\begin{aligned}
    \mathcal{E}^* = \left\{ e_i \,\middle|\,
    \text{Verifier}(x, y_i^{(\text{pe})}) > \text{Verifier}(x, y) \right. \\
    \left. e_i \in \mathcal{E}, \ \  y_i^{(\text{pe})} \in \mathcal{Y}_{\text{pe}} \right\} \subseteq \mathcal{E},
\end{aligned}
\end{equation}
where a new subset of errors $\mathcal{E}^*$ that contribute to quality improvement is identified, while non-impactful errors are discarded.

\subsection{Post-process: Error-Based Scoring} \label{sec:method_postprocess}

We adopt the MQM weighting scheme \citep{freitag-etal-2021-experts} for human scoring of LLM-generated errors, consistent with previous works \citep{lu-etal-2024-error, kocmi-federmann-2023-gemba, fernandes-etal-2023-devil}. The final score is calculated as the weighted sum of different error types:
\begin{equation}
\begin{split}
    \text{Score} = \max(-25, & -25 N_{\text{critical}} \\
    & - 5 N_{\text{major}} - N_{\text{minor}}),
\end{split}
\end{equation}
where $N_{\text{critical}}$, $N_{\text{major}}$, and $N_{\text{minor}}$ denote the number of critical, major, and minor errors, respectively. A lower bound of $-25$ is set to prevent the score from becoming excessively negative if too many errors are identified by the evaluator. 

Finally, our approach offers a comprehensive evaluation, including scores reflecting translation quality, error spans that contribute to improvement, and post-edited translations as a byproduct.

\section{Experimental Setup}

\subsection{Test Dataset}

\paragraph{WMT22} WMT22\footnote{\url{https://www.statmt.org/wmt22/}} metrics shared tasks \citep{freitag-etal-2022-results} includes English-German (En-De), English-Russian (En-Ru), and Chinese-English (Zh-En) across four domains: conversational, e-commerce, news, and social, with expert human annotations. This study evaluates 106,758 segments from 54 MT systems\footnote{We select WMT22 to align our conclusions with other studies and exclude datasets from previous years to prevent potential data contamination.}.

\paragraph{IndicMT} IndicMT \citep{sai-b-etal-2023-indicmt} is a low-resource translation test set with MQM annotations, translating from English to four Indian languages: Assamese, Maithili, Kannada, and Punjabi. We include 1,000 segments, aligned with \citet{singh-etal-2024-good}, to evaluate the low-resource generalizability of our approach.

{\ding{43}} See Appendix~\ref{appendix:testset} for details about test sets.

\subsection{Large Language Models}

To verify the model-agnostic capability of MQM-APE, we adopt 8 LLMs with various model architectures, scales, and research purposes\footnote{As recommended by \citet{kocmi-federmann-2023-gemba}, we exclude closed-source LLMs like GPT-series due to their potential performance fluctuations with updates, and challenges in result reproducibility.}. The LLMs used are open-source and available from Huggingface\footnote{\url{https://huggingface.co/}}, ensuring reproducibility and transparency.

\paragraph{General-purpose LLMs} We consider three series of open-source models, Llama 3, Mixture of Experts, and Qwen1.5. We test two instruction-tuned LLMs from the Llama 3 series \citep{dubey2024llama}, developed by Meta: 8b (\textbf{Llama3-8b-inst}) and 70b (\textbf{Llama3-70b-inst}), which are widely used in research. The Mixtral models \citep{jiang2024mixtral} are a sparse mixture of experts, differing in architecture from Llama. We use 8x7b-Instruct-v0.1 (\textbf{Mixtral-8x7b-inst}) and 8x22b-Instruct-v0.1 (\textbf{Mixtral-8x22b-inst}) for testing. Qwen1.5 series, an improved version of Qwen \citep{yang2024qwen2} from Alibaba Cloud, is pretrained on a larger Chinese corpus. We test 14b (\textbf{Qwen1.5-14b-chat}) and 72b (\textbf{Qwen1.5-72b-chat}) models. 

\paragraph{Translation-specific LLMs} We evaluate on Tower \citep{alves2024tower}, a multilingual LLM by Unbabel, specifically trained from Llama 2 \citep{touvron2023llama} for translation-related tasks. This model excels in error detection and post-editing. We use TowerInstruct-7B-v0.2 (\textbf{Tower-7b-inst}) and TowerInstruct-13B-v0.1 (\textbf{Tower-13b-inst}).

\subsection{Prompts} 

We use consistent prompts for all tested LLMs without additional optimization techniques.

\paragraph{Error Analysis Evaluator} We implement the state-of-the-art GEMBA-MQM \citep{kocmi-federmann-2023-gemba}, a three-shot reference-free evaluation strategy originated from \citet{lu-etal-2024-error}. This language-agnostic prompt requires no revision for different language pairs.

\paragraph{Automatic Post Editor} We use a straightforward prompt to enable LLMs to post-edit target translations based on error spans.

\paragraph{Pairwise Quality Verifier} LLMs are prompted to select the better translation between the post-edited text and the original. We verify twice to mitigate positional bias \citep{shi2024judging}. 

{\ding{43}} See Appendix~\ref{appendix:prompt_contexts} for prompt contexts.

\subsection{Meta Evaluation} 

\paragraph{Reliability} We assess how well evaluator judgments align with human-annotated MQM score. At the system level, we follow \citet{kocmi-etal-2021-ship} to use pairwise accuracy (\textbf{Acc.}), which measures the agreement between metric and human rankings\footnote{Since IndicMT does not provide system-level information, we omit system-level scores for this test set.}. At the segment level, we apply group-by-item pairwise accuracy (\textbf{Acc*}) with tie calibration \citet{deutsch-etal-2023-ties}, using the $\text{acc}_{eq}^*$ variant to compare metric with gold scores. For reproductivity, we use MTME tools\footnote{\url{https://github.com/google-research/mt-metrics-eval}} recommended by WMT.

\paragraph{Interpretability} Following \citet{fernandes-etal-2023-devil, huang-etal-2024-lost}, we use the span precision (SP) and the major precision (MP) to evaluate the quality of the error spans compared to human-annotated spans in MQM. For a set of error spans $\mathcal{E} = \{e_1, \cdots, e_N\}$, where $e_j = \{w_i, w_{i+1}, \cdots\}$ represents an error span, $\mathcal{P}(e_j) = \{i|w_i \in e_j \}, \ \ j=1, \cdots, N$ denotes the position of errors. We measure overlap using $\mathcal{P}(\mathcal{E}) = \bigcup_{j=1}^N \mathcal{P}(e_j)$. SR and MR are defined as follows:
\begin{equation}
    \text{SP} = \frac{\mathcal{P}(\mathcal{E})\cap\mathcal{P}(\hat{\mathcal{E}})}{P(\hat{\mathcal{E}})},
\end{equation}
\begin{equation}
    \text{MP} = \frac{\mathcal{P}(\mathcal{E}_{\text{maj}})\cap\mathcal{P}(\hat{\mathcal{E}}_{\text{maj}})}{P(\hat{\mathcal{E}}_{\text{maj}})},
\end{equation}
where $\mathcal{E}$ and $\hat{\mathcal{E}}$ denote the gold error spans from MQM and the predicted error spans from LLM, respectively. The subscript "maj" indicates that the subset includes critical and major errors, the most severe types of translation errors\footnote{Recall-based metrics are not used because MQM-APE extracts from original error spans without introducing new ones, making metrics like SR and MR unsuitable for evaluation.}.

\paragraph{Significance Analysis} At the segment level, we follow WMT22 metrics shared task to utilize PERM-BOTH hypothesis test \citep{deutsch-etal-2021-statistical} to assess the significance of metrics. We use 1000 re-sampling runs and set $p=0.05$. The same significance analysis is applied to error span quality. 

{\ding{43}} See in Table~\ref{tab:appendix_wmt_res} and Table~\ref{tab:appendix_error_res} in Appendix~\ref{appendix:detail_res}, where "\dag" indicates cases where MQM-APE significantly outperforms GEMBA-MQM on specific meta-evaluation metrics.

\subsection{Alignment with Human Judgments} 

\begin{table}[ht]
\centering
\small
\setlength{\tabcolsep}{6pt}
\begin{tabular}{cc}
\toprule[0.5mm]
\textbf{Estimated Accuracy $\geq 95\%$} & \textbf{$\Delta$} \\ \midrule
$\text{CometKiwi}_{22}^{\text{QE}}$ & 1.18 \\
$\text{BLEURT}_{20}$ & 2.44 \\\bottomrule[0.5mm]

\end{tabular}
\caption{\textbf{Thresholds of metrics} used in translation performance comparison. For instance, to achieve alignment with human judgments at 95\% confidence, the score improvement must be $\geq 1.18$ for $\text{CometKiwi}_{22}^{\text{QE}}$.}
\label{tab:metric_threshold}
\end{table}

Following \citet{kocmi-etal-2024-navigating}, we use the metric performance difference to assess alignment with human judgments. As shown in Table~\ref{tab:metric_threshold}, we present the threshold of metric delta for $\text{CometKiwi}_{22}^{\text{QE}}$ and $\text{BLEURT}_{20}$, which indicates $\geq$95\% confidence with human judgments. Detailed results on translations are reported in Table~\ref{tab:post-edit}.


\section{Results}

\subsection{Performance of MQM-APE} \label{sec:main_res}

\begin{table*}[ht]
\centering
\small
\setlength{\tabcolsep}{3pt}
\begin{tabular}{llccccccc}
\toprule[0.5mm]
\multirow{2}{*}{\textbf{Models}} & \multirow{2}{*}{\textbf{Strategy}} & \textbf{System-Level Acc.} & \multicolumn{4}{c}{\textbf{Segment-Level Acc*}} & \multicolumn{2}{c}{\textbf{Error span Quality}} \\
\cmidrule(lr){3-3} \cmidrule(lr){4-7} \cmidrule(lr){8-9}
 & & \textbf{All (3LPs)} & \textbf{En-De} & \textbf{En-Ru} & \textbf{Zh-En} & \textbf{Avg.} & \textbf{SP} & \textbf{MP} \\ \midrule
\multirow{2}{*}{\textbf{Llama3-8b-inst}} & \faToggleOff \ MQM & 77.4 & 54.5 & 48.3 & 45.7 &  49.5 & 9.0 & 4.9 \\
 & \faToggleOn \ MQM-APE & \makecell{83.2 \small\textcolor{red}{(+5.8)}} & 54.4 & 51.0 & 47.5 & \makecell{51.0 \small\textcolor{red}{(+1.5)} } & \makecell{10.4 \small\textcolor{red}{(+1.4)}} & \makecell{5.6 \small\textcolor{red}{(+0.7)}} \\ \midrule
\multirow{2}{*}{\textbf{Llama3-70b-inst}} & \faToggleOff \ MQM & 82.5 & 54.3 & 51.7 & 49.4 &  51.8 & 16.2 & 11.1 \\
 & \faToggleOn \ MQM-APE & \makecell{85.0 \small\textcolor{red}{(+2.5)}} & 55.7 & 53.3 & 50.8 & \makecell{53.3 \small\textcolor{red}{(+1.5)}} & 17.2 \textcolor{red}{(+1.0)} & 11.7 \textcolor{red}{(+0.6)} \\ \midrule
\multirow{2}{*}{\textbf{Mixtral-8x7b-inst}} & \faToggleOff \ MQM & 85.8 & 55.2 & 53.5 & 48.8 & 52.5 & 10.5 & 5.0 \\ 
 & \faToggleOn \ MQM-APE &  \makecell{85.8 \small(0.0) } & 55.4 & 53.7 & 50.2 &  \makecell{53.1 \small\textcolor{red}{(+0.6)}} & 10.7 \textcolor{red}{(+0.2)} & 5.1 \textcolor{red}{(+0.1)} \\ \midrule
\multirow{2}{*}{\textbf{Mixtral-8x22b-inst}} & \faToggleOff \ MQM &  87.2 & 55.7 & 53.4 & 50.3 &  53.1 & 9.9 & 6.7 \\
 & \faToggleOn \ MQM-APE &  \makecell{88.3 \small\textcolor{red}{(+1.1)}} & 56.9 & 55.1 & 50.6 &  \makecell{54.2 \small\textcolor{red}{(+1.1)}} & 10.3 \textcolor{red}{(+0.4)} & 6.9 \textcolor{red}{(+0.2)} \\ \midrule
\multirow{2}{*}{\textbf{Qwen1.5-14b-chat}} & \faToggleOff \ MQM &  83.9 & 55.6 & 51.0 & 48.2 &  51.6 & 9.8 & 4.6 \\
 & \faToggleOn \ MQM-APE &  \makecell{84.3 \small\textcolor{red}{(+0.4)}} & 55.7 & 51.9 & 49.8 &  \makecell{52.5 \small\textcolor{red}{(+0.9)}} & 10.3 \textcolor{red}{(+0.5)} & 4.8 \textcolor{red}{(+0.2)} \\ \midrule
\multirow{2}{*}{\textbf{Qwen1.5-72b-chat}} & \faToggleOff \ MQM &  84.7 & 56.0 & 54.7 & 50.6 &  53.8 & 9.2 & 3.3 \\ 
 & \faToggleOn \ MQM-APE &  \makecell{85.8 \small\textcolor{red}{(+1.1)}} & 56.4 & 55.7 & 51.4 &  \makecell{54.5 \small\textcolor{red}{(+0.7)}} & 10.3 \textcolor{red}{(+1.1)} & 3.7 \textcolor{red}{(+0.4)} \\ \midrule
\multirow{2}{*}{\textbf{Tower-7b-inst}} & \faToggleOff \ MQM &  81.8 & 53.6 & 49.6 & 42.1 & 48.4 & 17.2 & 4.1 \\
 & \faToggleOn \ MQM-APE &  \makecell{84.3 \small\textcolor{red}{(+2.5)}} & 53.5 & 50.9 & 44.4 & \makecell{49.6 \small\textcolor{red}{(+1.2)}} & 17.5 \textcolor{red}{(+0.3)} & 4.3 \textcolor{red}{(+0.2)} \\ \midrule
\multirow{2}{*}{\textbf{Tower-13b-inst}} & \faToggleOff \ MQM &  83.6 & 55.8 & 55.7 & 44.9 & 52.1 & 21.2 & 16.1 \\
 & \faToggleOn \ MQM-APE &  \makecell{85.0 \small\textcolor{red}{(+1.4)}} & 56.3 & 55.7 & 45.0 &  \makecell{52.3 \small\textcolor{red}{(+0.2)}} & 21.4 \textcolor{red}{(+0.2)} & 15.8 \textcolor{blue}{(-0.3)} \\ \bottomrule[0.5mm]
\end{tabular}
\caption{\textbf{Comparison of performance between GEMBA-MQM ("MQM") and MQM-APE} on WMT22 with human-labeled MQM, evaluated using pairwise accuracy (\%) at the system level, pairwise accuracy with tie calibration (\%) at the segment level, and error span precision of errors (SP) and major errors (MP), respectively.}
\label{tab:mainres}
\end{table*}

\begin{table}[ht]
\centering
\small
\setlength{\tabcolsep}{5pt}
\begin{tabular}{llc}
\toprule[0.5mm]
\textbf{Models} & \textbf{Prompt} & \textbf{SEG Acc*} \\\midrule
\multirow{2}{*}{\textbf{Llama3-8b-inst}} & \faToggleOff \ MQM & 34.1 \\
& \faToggleOn \ MQM-APE & \makecell{41.5 \small\textcolor{red}{(+7.4)}} \\\midrule
\multirow{2}{*}{\textbf{Llama3-70b-inst}} & \faToggleOff \ MQM & 38.7 \\
& \faToggleOn \ MQM-APE & \makecell{44.1 \small\textcolor{red}{(+5.4)}} \\\midrule
\multirow{2}{*}{\textbf{Mixtral-8x7b-inst}} & \faToggleOff \ MQM & 35.7 \\
& \faToggleOn \ MQM-APE & \makecell{40.5 \small\textcolor{red}{(+4.8)}} \\\midrule
\multirow{2}{*}{\textbf{Mixtral-8x22b-inst}} & \faToggleOff \ MQM & 44.0 \\
& \faToggleOn \ MQM-APE & \makecell{45.4 \small\textcolor{red}{(+1.4)}} \\\midrule
\multirow{2}{*}{\textbf{Qwen1.5-14b-chat}} & \faToggleOff \ MQM & 23.0 \\
& \faToggleOn \ MQM-APE & \makecell{37.4 \small\textcolor{red}{(+14.4)}} \\\midrule
\multirow{2}{*}{\textbf{Qwen1.5-72b-chat}} & \faToggleOff \ MQM & 43.9 \\
& \faToggleOn \ MQM-APE & \makecell{44.9 \small\textcolor{red}{(+1.0)}} \\\midrule
\multirow{2}{*}{\textbf{Tower-7b-inst}} & \faToggleOff \ MQM & 26.0 \\
& \faToggleOn \ MQM-APE & \makecell{33.2 \small\textcolor{red}{(+7.2)}} \\\midrule
\multirow{2}{*}{\textbf{Tower-13b-inst}} & \faToggleOff \ MQM & 34.0 \\
& \faToggleOn \ MQM-APE & \makecell{34.5 \small\textcolor{red}{(+0.5)} \ } \\
\bottomrule[0.5mm]
\end{tabular}
\caption{\textbf{Segment-level comparison between GEMBA-MQM ("MQM") and MQM-APE} on IndicMT.}
\label{tab:indicmt_res}
\end{table}

We evaluate our proposed MQM-APE against GEMBA-MQM ("MQM") across different LLMs, with the results presented in Table~\ref{tab:mainres} for WMT22 and Table~\ref{tab:indicmt_res} for IndicMT. The results indicate that:

{\ding{43}} See detailed results in Appendix~\ref{appendix:detail_res} for a comprehensive view of the performance. 

\paragraph{(i) Reliability: MQM-APE consistently enhances GEMBA-MQM at both the system and segment levels for all tested LLMs.}

Building on prior findings that MQM-based evaluators like GEMBA-MQM achieve state-of-the-art performance at the system level~\citep{kocmi-federmann-2023-gemba, lu-etal-2024-error}, we show that our MQM-APE approach consistently improves performance across all three language pairs. At the segment level, MQM-APE also surpasses the performance against GEMBA-MQM, indicating better reliability for LLM-based translation evaluators.

Notably, this improvement is observed across all LLMs tested, except for Mixtral-8x22b-inst, which maintains the same performance at the system level, but improves at the segment level.

\paragraph{(ii) Interpretability: MQM-APE obtains better error span quality compared with GEMBA-MQM.}

With explainable metrics emerging as a promising direction \citep{xu-etal-2023-instructscore, leiter-etal-2023-eval4nlp}, we find that our approach enhances the quality of predicted error spans compared to human annotations, with SP increasing across all tested LLMs, and MP improving in 7 out of 8. This suggests that MQM-APE helps LLMs identify high-quality errors and provide fine-grained feedback.

\paragraph{(iii) Evaluator Applicability: MQM-APE complements LLM-based evaluators specifically trained for translation-related tasks.}

For translation-specific LLM-based evaluators like Tower \citep{alves2024tower}, MQM-APE demonstrates broad applicability, enhancing performance at both system and segment levels, and obtaining better quality of error spans. This suggests that MQM-APE complements various pretraining or tuning strategies. However, MQM-APE has less impact on major error span precision (MP) for Tower-13b-inst, potentially because it sometimes corrects multiple major errors simultaneously, limiting the effectiveness of error discrimination. A potential solution is to apply APE to minor errors only, balancing reliability and interpretability.

\paragraph{(iv) Language Generalizability: MQM-APE shows consistent improvements on low-resource test sets.}

We evaluated the generalizability of our approach using IndicMT, which includes human annotations for four low-resource Indian languages. Consistent with high-resource scenarios in WMT, MQM-APE significantly improves GEMBA-MQM across all LLMs, demonstrating enhanced reliability on four low-resource languages.

\subsection{Automatic Post-Editor} \label{sec:res_postedit}

\begin{table*}[ht]
\centering
\small
\setlength{\tabcolsep}{3pt}
\begin{tabular}{lcccccccc}
\toprule[0.5mm]
 \multirow{2}{*}{\textbf{Models}} & \multicolumn{3}{c}{$\textbf{CometKiwi}_{22}^{\text{QE}}$} & \multicolumn{3}{c}{$\textbf{BLEURT}_{20}$} & \multicolumn{2}{c}{\textbf{Segment Comparison Rates: APE vs. TGT}} \\
 \cmidrule(lr){2-4} \cmidrule(lr){5-7} \cmidrule(lr){8-9}
 & \textbf{TGT} & \textbf{APE} & \textbf{$\Delta$} & \textbf{TGT} & \textbf{APE} & \textbf{$\Delta$} & \textbf{\colorbox{mylightblue}{Win}/\colorbox{mylightyellow}{Tie}/\colorbox{mylightpink}{Lose}} & \textbf{Win lose ratio} \\ \midrule
\textbf{Llama3-8b-inst} & 77.68 & 79.14 & \cellcolor{myblue}+1.46\dag & 69.66 & 71.16 & \cellcolor{myblue}+1.50 & \hbarthree{46}{32}{22} & \cellcolor{myblue}2.10 \\
\textbf{Llama3-70b-inst} & 77.06 & 79.75 & \cellcolor{myblue}+2.69\dag & 68.49 & 71.63 & \cellcolor{myblue}+3.14\dag & \hbarthree{56}{30}{14} & \cellcolor{myblue}4.09 \\ \midrule
\textbf{Mixtral-8x7b-inst} & 75.46 & 77.59 & \cellcolor{myblue}+2.14\dag & 66.34 & 68.71 & \cellcolor{myblue}+2.37 &  \hbarthree{60}{24}{16} & \cellcolor{myblue}3.64 \\
\textbf{Mixtral-8x22b-inst} & 78.65 & 81.60 & \cellcolor{myblue}+2.96\dag & 70.37 & 74.39 & \cellcolor{myblue}+4.02\dag & \hbarthree{62}{28}{10} & \cellcolor{myblue}6.55 \\ \midrule
\textbf{Qwen1.5-14b-chat} & 72.82 & 77.41 & \cellcolor{myblue}+4.60\dag & 62.35 & 67.71 & \cellcolor{myblue}+5.36\dag & \hbarthree{64}{24}{12} & \cellcolor{myblue}5.56 \\ 
\textbf{Qwen1.5-72b-chat} & 76.82 & 79.87 & \cellcolor{myblue}+3.05\dag & 68.17 & 71.79 & \cellcolor{myblue}+3.62\dag & \hbarthree{55}{30}{15} & \cellcolor{myblue}3.78 \\ \midrule
\textbf{Tower-7b-inst} & 74.43 & 75.02 & \cellcolor{myblue}+0.59 & 64.47 & 66.19 & \cellcolor{myblue}+1.72 & \hbarthree{40}{30}{30} & \cellcolor{myblue}1.31 \\
\textbf{Tower-13b-inst} & 75.32 & 78.85 & \cellcolor{myblue}+3.54\dag & 64.90 & 71.01 & \cellcolor{myblue}+6.12\dag & \hbarthree{63}{28}{9} & \cellcolor{myblue}7.35 \\
\bottomrule[0.5mm]
\end{tabular}
\caption{\textbf{Performance of Automatic Post Editor} measured with $\text{CometKiwi}_{22}^{\text{QE}}$ and $\text{BLEURT}_{20}$. "\dag" indicates that the metrics difference ($\Delta$) has >95\% estimated accuracy with humans \citep{kocmi-etal-2024-navigating}. For segment comparison, we define \colorbox{mylightblue}{Win} as cases where both $\text{CometKiwi}_{22}^{\text{QE}}$ and $\text{BLEURT}_{20}$ rate APE higher than TGT, \colorbox{mylightpink}{Lose} where they rate APE lower, and \colorbox{mylightyellow}{Tie} when their evaluations conflict.}
\label{tab:post-edit}
\end{table*}

We evaluate the effectiveness of APE modules by comparing the overall quality between the original translation ("TGT") and the post-edited translation ("APE") using $\text{CometKiwi}{22}^{\text{QE}}$ and $\text{BLEURT}_{20}$, as recommended by \citet{kocmi-etal-2024-navigating}. $\text{CometKiwi}_{22}^{\text{QE}}$, a reference-free metric, is highly discriminative of system quality, while $\text{BLEURT}_{20}$, a reference-based metric, offers complementary insights, enhancing the reliability of our findings.

Table~\ref{tab:post-edit} shows performance comparisons\footnote{Note that only APEs related to impactful errors are considered in this analysis, as not all errors contribute to the effectiveness in improving translation quality.}. All tested LLMs show improved APE translation quality over the original translation ("TGT") on both $\text{CometKiwi}_{22}^{\text{QE}}$ and $\text{BLEURT}_{20}$. Additionally, 7 out of 8 LLMs achieve >95\% estimated accuracy (marked with "$\dag$") on metric delta, reflecting high confidence related to human judgments in the observed performance improvements. Furthermore, we observe that the “Win lose ratio” exceeds 1 for all tested LLMs, indicating that APE outperforms original translation in pairwise segment comparisons. This again confirms APE's effectiveness.

\subsection{Pairwise Quality Verifier} \label{sec:res_verifier}

\begin{table}[ht]
\centering
\small
\setlength{\tabcolsep}{5pt}
\begin{tabular}{lcccccc}
\toprule[0.5mm]
 \multirow{2}{*}{\textbf{Models}} & \multicolumn{3}{c}{$\textbf{CometKiwi}_{22}^{\text{QE}}$} & \multicolumn{3}{c}{$\textbf{BLEURT}_{20}$} \\
 \cmidrule(lr){2-4} \cmidrule(lr){5-7}
 & \textbf{P} & \textbf{R} & \textbf{F1} & \textbf{P} & \textbf{R} & \textbf{F1} \\ \midrule
\textbf{Llama3-8b-inst}     & 64 & 84 & 72 & 60 & 80 & 68 \\
\textbf{Llama3-70b-inst}    & 74 & 94 & 83 & 69 & 93 & 79 \\ \midrule
\textbf{Mixtral-8x7b-inst}  & 66 & 93 & 77 & 61 & 93 & 73 \\
\textbf{Mixtral-8x22b-inst} & 76 & 94 & 84 & 71 & 95 & 81 \\ \midrule
\textbf{Qwen1.5-14b-chat}    & 71 & 99 & 83 & 69 & 98 & 81 \\
\textbf{Qwen1.5-72b-chat}    & 70 & 95 & 80 & 66 & 93 & 78 \\ \midrule
\textbf{Tower-7b-inst}      & 57 & 65 & 61 & 54 & 82 & 65 \\
\textbf{Tower-13b-inst}     & 80 & 93 & 86 & 82 & 92 & 86 \\
\bottomrule[0.5mm]
\end{tabular}
\caption{\textbf{Comparison of the pairwise quality verifier's consistency} with $\text{CometKiwi}_{22}^{\text{QE}}$ and $\text{BLEURT}_{20}$, which serve as ground truth.}
\label{tab:verifier}
\end{table}

Table~\ref{tab:verifier} compares the consistency of pairwise judgments from the verifier with numerical metrics, $\text{CometKiwi}_{22}^{\text{QE}}$ and $\text{BLEURT}_{20}$, which serve as ground truth. Most LLMs achieve over 90\% recall ("R") for $\text{CometKiwi}_{22}^{\text{QE}}$ and over 80\% for $\text{BLEURT}_{20}$, showing that the verifier aligns well with these metrics. However, the slightly lower precision ("P") suggests that the verifier is somewhat more lenient. Interestingly, Tower-7b-inst is less effective as an APE or quality verifier, while Tower-13b-inst excels in both roles.

\section{Analysis}

\subsection{MQM-APE Exhibits Superior Performance Compared to Random Error Filter} \label{sec:random}

\begin{figure}[ht]

\includegraphics[scale=0.29]{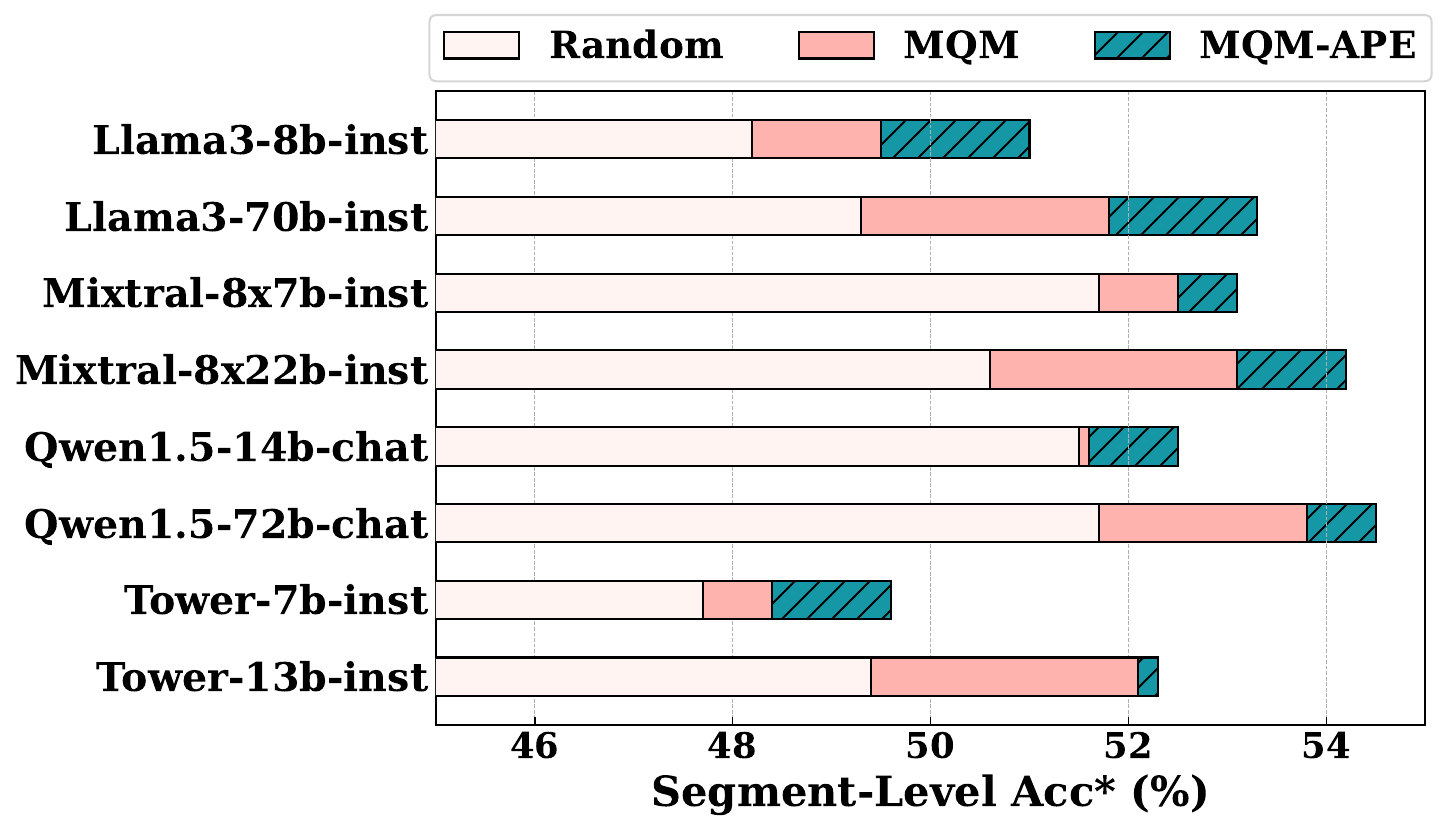}
\centering
\caption{\textbf{Comparison between MQM-APE, random error filter ("Random")} and  GEMBA-MQM ("MQM") on segment-level performance.}
\label{fig:random_compare}
\end{figure}

A potential concern with MQM-APE is that performance improvements might be attributed to the fewer identified errors. Figure~\ref{fig:random_compare} compares MQM-APE with a trivial random error filter—where errors are randomly discarded rather than based on the quality difference in APE. In contrast to MQM-APE, which enhances performance against GEMBA-MQM across all tested LLMs, the random filter consistently degrades performance, highlighting the importance of APE for error extraction.

\subsection{MQM-APE Introduces an Acceptable Inference Cost Compared to the Original} \label{sec:inference_cost}

\begin{table}[ht]
\centering
\small
\setlength{\tabcolsep}{3pt}
\begin{tabular}{cccc}
\toprule[0.5mm]
\multirow{2}{*}{ \textbf{LLM Module}} & \multirow{2}{*}{\textbf{Extra?}} & \multicolumn{2}{c}{\textbf{No. of Tokens}} \\
\cmidrule(lr){3-4}
 & & \textbf{Input} & \textbf{Generated} \\ \midrule
 \makecell{Error Analysis Evaluator\\(GEMBA-MQM)} & - & 1295.65 & 46.70 \\\midrule
 Error-based APE & \textcolor{red}{+} & 265.88 & 96.29 \\\midrule
 Pairwise Quality Verifier & \textcolor{red}{+} & 349.46 & 12.68 \\
\bottomrule[0.5mm]
\end{tabular}
\caption{\textbf{Average number of input and generated tokens} per segment for each module. "\textcolor{red}{+}" indicates the additional modules introduced in MQM-APE.}
\label{tab:inference}
\end{table}

Table~\ref{tab:inference} presents the input and inference costs for each LLM during inference, analyzing the additional cost associated with MQM-APE. Specifically, the extra cost arises from the APE and quality verifier, which generate approximately twice as many tokens as the evaluator, while the input tokens are about half as many. 

{\ding{43}} See Appendix~\ref{appendix:infer_cost} for a detailed analysis on MQM-APE, since inference costs vary depending on translation quality and the LLMs used.

\subsection{MQM-APE Can Replace the Quality Verifier with Metrics for Comparable Performance} \label{sec:replace_metrics}

\begin{figure}[ht]
\includegraphics[scale=0.29]{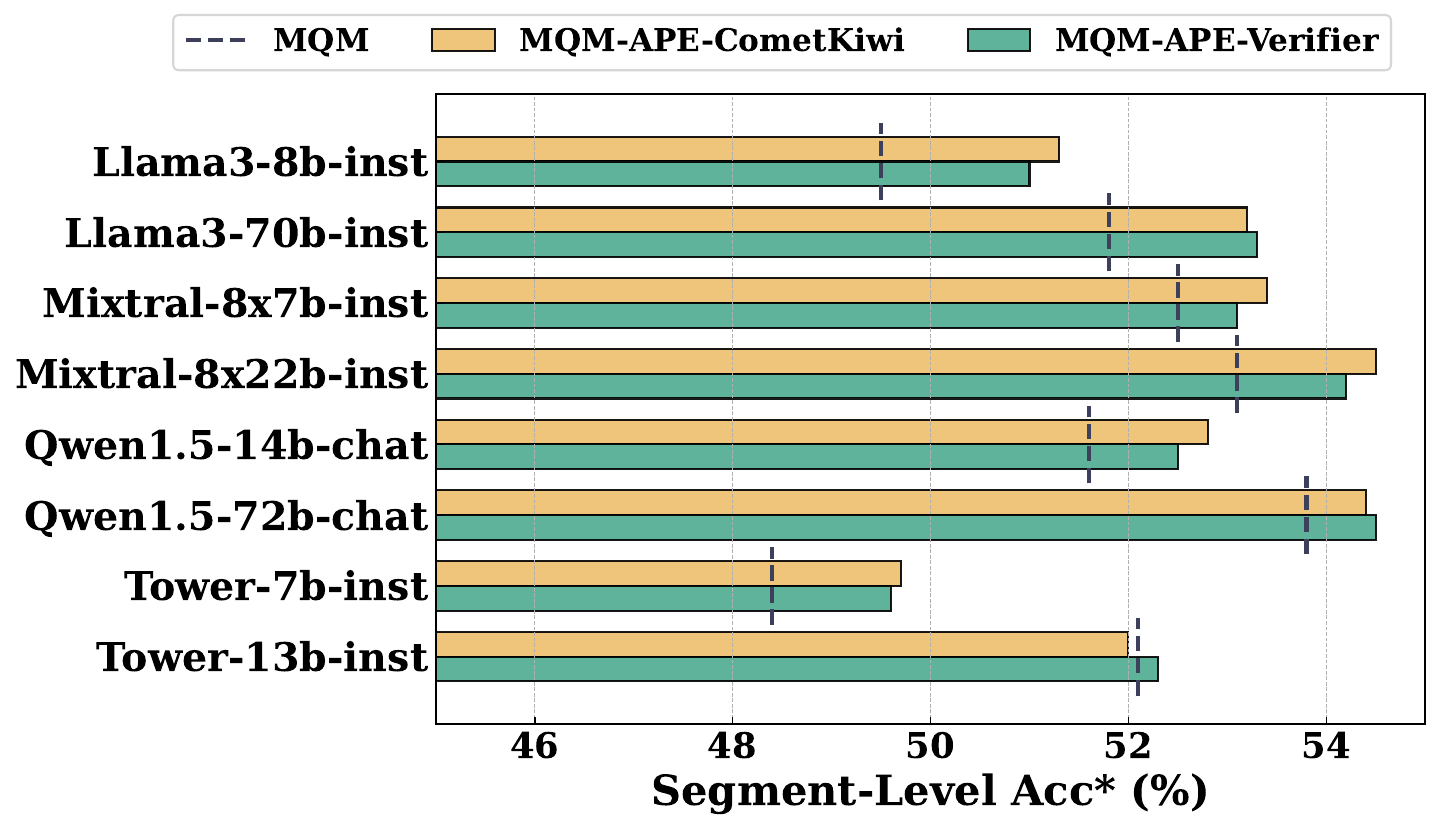}
\centering
\caption{\textbf{Comparison between MQM-APE with an LLM verifier and with $\textbf{CometKiwi}_{\textbf{22}}^{\textbf{QE}}$} as a replacement on segment-level performance.}
\label{fig:metric_compare}
\end{figure}

A cost-reducing alternative of MQM-APE is to replace the verifier with metrics to perform comparisons. Figure~\ref{fig:metric_compare} compares the performance of using either an LLM verifier or $\text{CometKiwi}_{22}^{\text{QE}}$\footnote{We adopt $\text{CometKiwi}_{22}^{\text{QE}}$ to maintain compatibility in reference-free evaluation.}. Both approaches achieve comparable effects, consistently surpassing GEMBA-MQM for most LLMs.




\subsection{MQM-APE Preserves Error Distribution Across Severities and Categories} \label{sec:error_dist}

\begin{figure}[ht]
\includegraphics[scale=0.29]{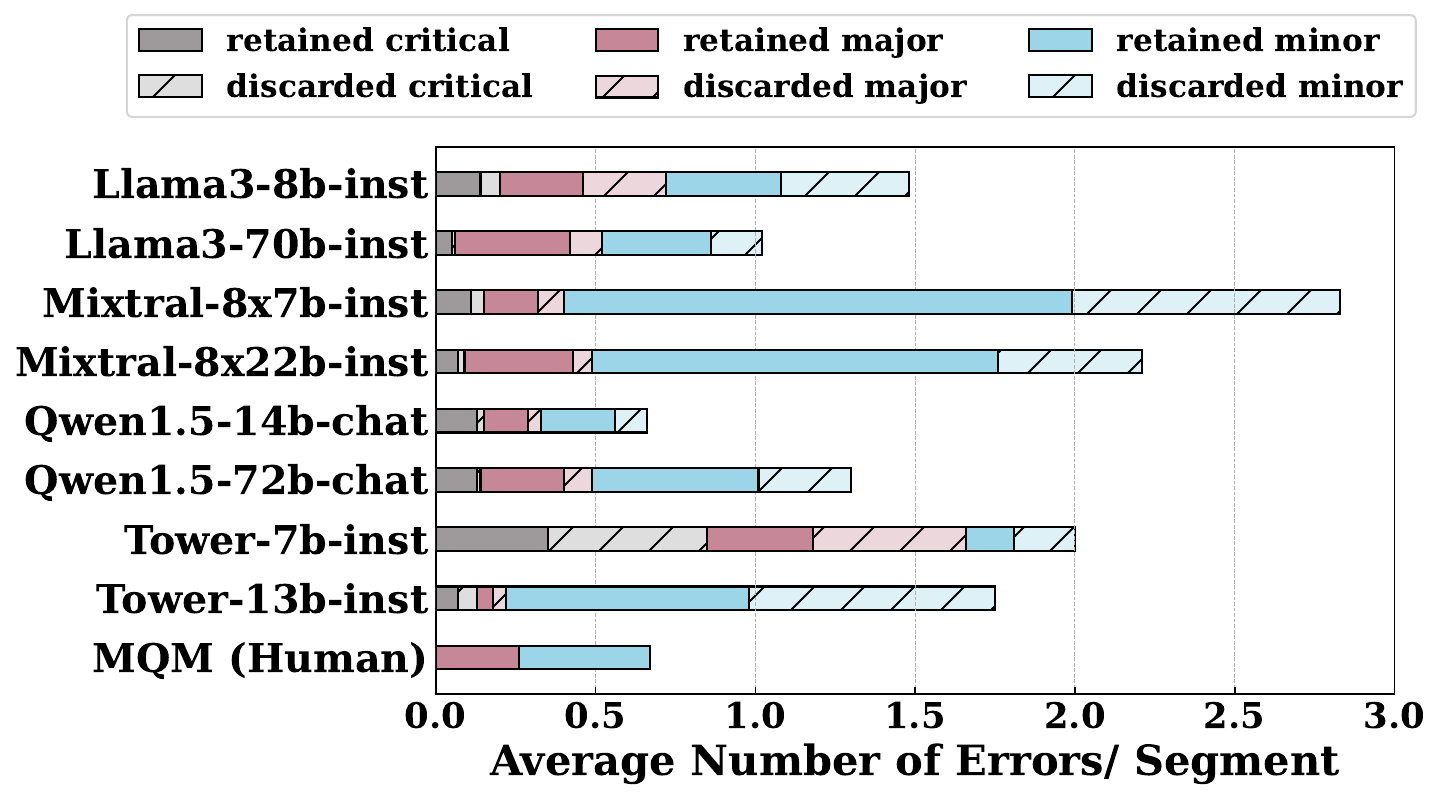}
\centering
\caption{\textbf{Average number of errors retained or discarded} for each severity level with MQM-APE.}
\label{fig:errors_compare}
\end{figure}

\begin{figure}[ht]
\includegraphics[scale=0.29]{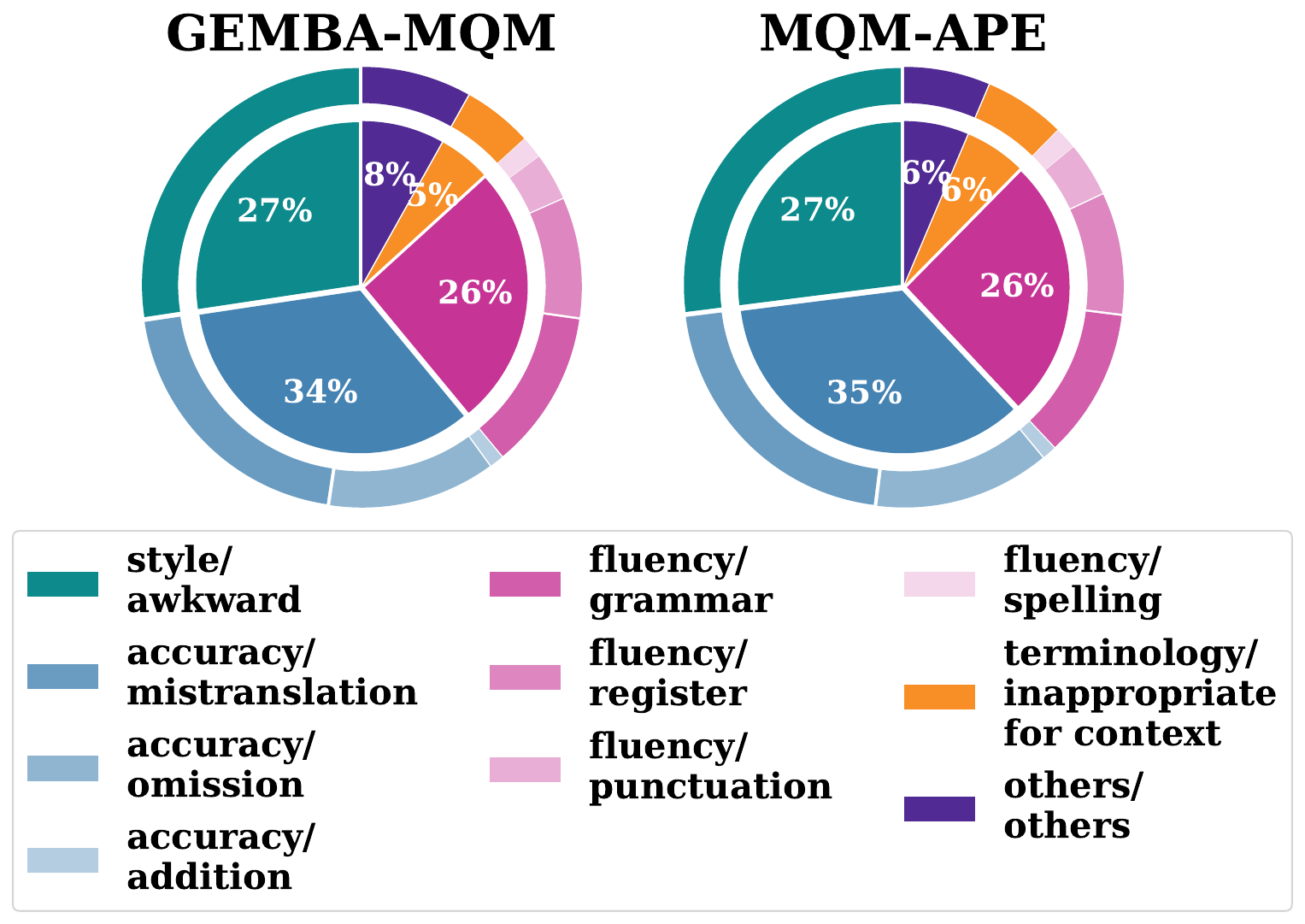}
\centering
\caption{\textbf{Distribution of error categories} between GEMBA-MQM and MQM-APE.}
\label{fig:category_compare}
\end{figure}

Figure~\ref{fig:errors_compare} shows the influence on the number of errors for each severity level. Overall, MQM-APE retains a similar error distribution compared to GEMBA-MQM. Discarded errors come mainly from minor ones, with minimal changes to critical or major errors, which aligns with our expectation that more severe errors have a greater impact on quality improvement. Next, we compare the error categories between GEMBA-MQM and MQM-APE, as shown in Figure~\ref{fig:category_compare}. The categories of errors remain largely consistent after MQM-APE. 

{\ding{43}} See Appendix~\ref{appendix:error_analysis} for a detailed analysis across error severities and categories, which further discusses their alignment with human annotations.

\subsection{Recommendation on LLM Selection when using LLM-based Evaluators} \label{sec:recommend}

\begin{table*}[ht]
\centering
\small
\setlength{\tabcolsep}{6pt}
\begin{tabular}{lcccc}
\toprule[0.5mm]
\multirow{2}{*}{\textbf{Models}} & \multirow{2}{*}{\textbf{Scale}} & \textbf{Reliability} & \textbf{Interpretability} & \textbf{Inference Cost} \\
& & SYS Acc. & Span Precision & \#Token Generated \\\midrule
\textbf{Llama3-8b-inst} & \ding{83} Small & \fiv{83.2} & \fou{10.4} & \thi{113.4} \\
\textbf{Llama3-70b-inst} & \ding{107} Large & \thi{85.0} & \two{17.2} & \two{79.6} \\
\textbf{Mixtral-8x7b-inst} & \ding{107} Large & \two{85.8} & \thi{10.7} & \six{339.3} \\
\textbf{Mixtral-8x22b-inst} & \ding{107} Large & \win{88.3} & \fou{10.3} & \fiv{211.5} \\
\textbf{Qwen1.5-14b-chat} & \ding{83} Small & \fou{84.3} & \fou{10.3} & \win{56.7} \\ 
\textbf{Qwen1.5-72b-chat} & \ding{107} Large & \two{85.8} & \fou{10.3} & \thi{93.3} \\
\textbf{Tower-7b-inst} & \ding{83} Small & \fou{84.3} & \two{17.5} & \fou{150.0} \\
\textbf{Tower-13b-inst} & \ding{83} Small & \thi{85.0} & \win{21.4} & \fiv{201.6} \\
\bottomrule[0.5mm]
\end{tabular}
\caption{\textbf{Comparison of different LLMs} for MQM-APE, with their performance rankings across various aspects.}
\label{tab:llmselection}
\end{table*}


We present a performance ranking of different backbone LLMs as translation evaluators in Table~\ref{tab:llmselection}, considering three aspects: reliability, interpretability, and inference cost. This analysis offers a comprehensive guide for selecting LLMs when implementing MQM-APE. For instance, while Mixtral-8x22b-inst may be the most reliable evaluator, it could produce inaccurate and redundant error spans, along with higher inference costs. In practice, users can choose the most appropriate LLM based on available computational resources and the desired trade-offs between reliability, interpretability, and inference cost for quality assessment.

\section{Related Work}

\paragraph{MT Metrics and Error Annotation} Evaluation metrics are of crucial importance to the development of MT systems \citep{freitag-etal-2022-results}. Since traditional metrics such as BLEU \citep{papineni-etal-2002-bleu} are unreliable for evaluating high-quality MT systems \citep{mathur-etal-2020-tangled}, metrics such as COMET-Kiwi \citep{rei-etal-2022-cometkiwi} and BLEURT \citep{sellam-etal-2020-bleurt} have succeeded in aligning with human. As the demand for interpretability grows \citep{lu-etal-2023-toward}, we enhance the quality of error annotations by incorporating APE into evaluators.

\paragraph{LLM-based Evaluators} Leveraging LLMs as evaluators has become a prevalent approach \citep{zheng2024judging,liu-etal-2023-g}. GEMBA \citep{kocmi-federmann-2023-large} pioneered the LLM translation evaluator via direct prompting. Error Analysis Prompting \citep{lu-etal-2024-error}, which integrates Chain-of-Thought \citep{wei2022chain} to prompt LLMs in detecting explicit errors, is further enhanced by AutoMQM \citep{fernandes-etal-2023-devil} and GEMBA-MQM \citep{kocmi-federmann-2023-gemba}. Another line of research \citep{xu-etal-2023-instructscore, guerreiro2023xcomet, treviso2024xtower} fine-tunes LLMs for accurate error span prediction. We integrate APE into LLM-based evaluators, propose a training framework with consistent improvements.

\paragraph{APE and Self-Correction Techniques} APE has shown quality improvements and reduced translationese before the LLM era \citep{freitag-etal-2019-ape,chatterjee-etal-2020-findings}. LLMs, especially GPT-4, have demonstrated potential for identifying intentions~\cite{zhang2024intention}, performing multi-step advanced reasoning~\cite{zhong2024achieving}, and correcting errors ~\citep{raunak-etal-2023-leveraging}. Recently, \citet{ki-carpuat-2024-guiding} observes better translation quality with fine-grained annotations. Similarly, studies on self-correction \citep{madaan2024self} enhance translation through iterative corrections \citep{chen2023iterative, pan-etal-2024-automatically}, with \citet{xu-etal-2024-llmrefine} proposing search techniques for feedback. Orthogonal to their works, we leverage APE to obtain better evaluation with higher error span quality.

\section{Conclusion}
In this work, we introduce the MQM-APE framework, which incorporates APE for filtering out non-impactful errors that do not contribute to quality improvement. Experiments show consistent performance enhancements and better quality of error annotations across various LLMs and high- and low-resource language pairs. Future work will explore whether different LLMs can collaborate to achieve better results or whether varying prompting strategies can enhance predicted error quality.

\section*{Limitations}

\paragraph{Extra Inference Cost.} We acknowledge that MQM-APE incurs additional costs compared to standard MQM-like error analysis prompting. As discussed in \S\ref{sec:inference_cost}, MQM-APE requires approximately twice the number of tokens for inference. To mitigate this drawback, we propose a cost-reducing alternative by replacing the pairwise verifier with numerical metrics (\S\ref{sec:replace_metrics}), along with recommendations for LLM selection under budget constraints (\S\ref{sec:recommend}). Given MQM-APE’s improved reliability, high-quality error spans, and superior APE translations, we consider the extra cost both valuable and acceptable.

\paragraph{Error Distribution.} As discussed in \S\ref{sec:error_dist} and in Appendix~\ref{appendix:error_analysis}, while MQM-APE filters non-impactful errors and improves reliability and interpretability, it does not align the error distribution with human evaluation. We acknowledge that this distribution alignment issue warrants further investigation, and we leave it for future research.

\paragraph{Single LLM considered.} In our experiments, we use the single LLM for error analysis, APE, and quality verifier to support our main claim that MQM-APE improves reliability and interpretability for LLM evaluators. To ensure fair comparison and verification, we apply the same LLM without fine-tuning across different tasks. However, in real-world evaluations, multiple LLMs can collaborate for more robust assessments. Future work could explore these approaches.

\paragraph{Possibly Invalid Response.} Invalid or rejected responses introduce noise into the assessment process. This may result from the LLM's insufficient instruction-following capabilities, and models frequently exhibiting this behavior should be excluded from evaluations. We analyze this phenomenon in Appendix~\ref{appendix:bias_and_invalid}. Fortunately, such cases are rare in our experiments and can be considered negligible. Following \citet{kocmi-federmann-2023-large}, we will regenerate responses by slightly increasing the temperature if this happens.

\section*{Ethics Statement}

We prioritize ethical considerations and strictly comply with the Code of Ethics. All procedures in this study align with ethical standards. This paper centers on generating high-quality error spans by integrating APE into LLM translation evaluation. Our approach, MQM-APE, avoids inducing the model to generate harmful content. It only extracts error spans from the model's responses, reducing potential risks. The datasets and models employed are publicly available and commonly used in research. Since our model does not require fine-tuning, it minimizes the risks associated with learning from user inputs and avoids posing threats to the NLP community. We ensure that the findings are reproducible and reported accurately and objectively.

\section*{Acknowledgments}

We thank the anonymous reviewers and the area chair for their insightful comments and suggestions. This research is supported by the National Natural Science Foundation of China under Grant 61973083, the Shenzhen Science and Technology Program JCYJ20210324121213036, the RIE2025 Industry Alignment Fund – Industry Collaboration Projects (IAF-ICP) (Award I2301E0026), administered by A*STAR, as well as supported by Alibaba Group and NTU Singapore through Alibaba-NTU Global e-Sustainability CorpLab (ANGEL).

\bibliography{coling_latex}

\appendix

\section{Description of MQM}
\label{appendix:MQM}

\begin{table}[ht]
\centering
\scriptsize
\begin{tabular}{ll}
\toprule[0.5mm] 
\textbf{System} &  Online-A.en\\
\textbf{Domain} &  conversational\\
\textbf{Doc\_id}  & 1 \\
\textbf{Seg\_id}  & 6 \\\midrule
\textbf{Source(Zh)} & \begin{CJK}{UTF8}{gbsn}请问，订单情况现在是什么样？\end{CJK} \\
\textbf{Reference(En)} & May I ask what the status of the order is now? \\
\textbf{Translation(En)} & Please ask, what is the order situation now? \\\midrule
\textbf{Major Error(s)} & "Please ask" - Accuracy/Mistranslation\\
\textbf{Minor Error(s)} & "situation" - Style/Awkward \\
\bottomrule[0.5mm]
\end{tabular}
\caption{\textbf{An example of MQM}, comprising information of the test sample along with human-annotated errors.}
\label{tab:appendix_mqmcase}
\end{table}

\paragraph{Multidimensional Quality Metric (MQM)} is a human evaluation framework \citep{Lommel2018} commonly used in WMT metrics shared tasks as the golden standard \citep{freitag-etal-2021-experts}. It is developed to evaluate and categorize errors in translations. The annotations from human experts are open-sourced and available from WMT22 metrics shared tasks for scientific research \citep{freitag-etal-2022-results}. Table~\ref{tab:appendix_mqmcase} shows an example annotated through MQM framework.

\paragraph{About annotators quality.} In WMT22, MQM annotations for En-De and Zh-En were sponsored and executed by Google, using 11 professional translators (7 for En-De, 4 for Zh-En). The annotations for En-Ru were provided by Unbabel who utilized 4 professional, native language annotators with ample translation experience. They have access to the full document context. 

\paragraph{About inter-rater agreement.} In MQM, each segment is annotated by 2 or 3 annotators. The final segment-level score is an average over scores from all annotators. As depicted in \citep{freitag-etal-2021-experts}, the pairwise inter-rater agreement is about 0.584 for En-De, and 0.412 for Zh-En, which is significantly better than other evaluation protocols such as Scalar Quality Metric and Direct Assessment.

\section{Description of the Test Sets} \label{appendix:testset}

\begin{table*}[ht]
\centering
\begin{tabular}{ccccc}
\toprule[0.5mm]
\textbf{Test Set} & \textbf{Language Pairs} & \textbf{Segments} & \textbf{Systems} & \textbf{Domains} \\\midrule
\multirow{3}{*}{\makecell{WMT22\\\citep{freitag-etal-2022-results}}} & En-De & 2037 & 17 & \multirow{3}{*}{\makecell{news, conversational,\\e-commerce, social}} \\
 & En-Ru & 2037 & 17 & \\
 & Zh-En & 1875 & 20 & \\\midrule
\multirow{4}{*}{\makecell{IndicMT\\\citep{singh-etal-2024-good}}} 
& En-As & 250 & - & \multirow{4}{*}{news, education, travel} \\
& En-Mai & 250 & - &  \\
& En-Kn & 250 & - &  \\
& En-Pa & 250 & - &  \\
\bottomrule[0.5mm]
\end{tabular}
\caption{\textbf{Statistics of testset}. Source and translations are from the WMT22 metrics shared task and low-resource MQM annotations are from IndicMT dataset. }
\label{tab:appendix_testset}
\end{table*}

We utilize the WMT22 shared task test set \citep{freitag-etal-2022-results}, which covers English-German (En-De), English-Russian (En-Ru), and Chinese-English (Zh-En) translations across four distinct domains: conversational, e-commerce, news, and social media. In total, this study evaluates 106,758 segments from 54 MT systems.

To assess the low-resource generalization capability of our approach, we follow \citep{singh-etal-2024-good} by using the IndicMT test set for four Indian languages—Assamese, Maithili, Kannada, and Punjabi—each with 250 segments, amounting to 1,000 segments in total. This test set is sampled from FLORES-101 dataset \citep{goyal-etal-2022-flores} and annotated using human evaluations based on the MQM framework \citep{sai-b-etal-2023-indicmt}.

Table~\ref{tab:appendix_testset} provides detailed statistics about our test set.




\section{Prompt Contexts} \label{appendix:prompt_contexts}

We present the three prompt contexts used in this work. These contexts are consistent across all LLMs and language pairs, as they are both language-agnostic and model-agnostic. Figure~\ref{fig:prompt} illustrates the prompt contexts applied in our experiments.

\paragraph{Error Analysis Evaluator} We utilize GEMBA-MQM \citep{kocmi-federmann-2023-gemba}, a fixed three-shot prompting technique for marking error quality spans. This approach is language-agnostic and does not require human references.

\paragraph{Automatic Post-Editor} A straightforward zero-shot prompting strategy is employed for post-editing.

\paragraph{Pairwise Quality Verifier} We implement a simple one-pass prompting strategy, where LLMs are prompted to select the better translation from two options.

\begin{figure*}[htb]
{\footnotesize
    \begin{Verbatim}[commandchars=+\[\]]
    
    +textbf[Error Analysis Evaluator:]
    
    (SYSTEM) You are an annotator for the quality of machine translation. Your task is to identify 
    errors and assess the quality of the translation.
    
    (USER) +textbf[{source_language}] source:\n
    ```+textbf[{source_segment}]```\n
    +textbf[{target_language}] translation:\n
    ```+textbf[{target_segment}]```\n
    \n
    Based on the source segment and machine translation surrounded with triple backticks, identify
    error types in the translation and classify them. The categories of errors are: accuracy 
    (addition, mistranslation, omission, untranslated text), fluency (character encoding, grammar, 
    inconsistency, punctuation, register, spelling), style (awkward), terminology (inappropriate 
    for context, inconsistent use), non-translation, other, or no-error.\nEach error is classified
    as one of three categories: critical, major, and minor. Critical errors inhibit comprehension
    of the text. Major errors disrupt the flow, but what the text is trying to say is still
    understandable. Minor errors are technically errors, but do not disrupt the flow or hinder
    comprehension.

    (ASSISTANT) +textbf[{observed error classes}]
    
    \end{Verbatim}
}
{\footnotesize
    \begin{Verbatim}[commandchars=+\[\]]
    +textbf[Automatic Post-Editor:]
    
    (USER) +textbf[{source_language}] source: "+textbf[{source_segment}]"\n
    +textbf[{target_language}] translation: "+textbf[{target_segment}]"\n
    \n
    Please post-edit the translation to address the identified error: "+textbf[{error_category}] - 
    +textbf[{error_content}]". Provide only the corrected +textbf[{target_language}] translation after "Corrected 
    Translation:" without adding any additional explanations or translation information.

    (ASSISTANT) +textbf[{post-edited translation}]
    
    \end{Verbatim}
}
{\footnotesize
    \begin{Verbatim}[commandchars=+\[\]]
    +textbf[Pairwise Quality Verifier:]
    
    (USER) +textbf[{source_language}] source: "+textbf[{source_segment}]"\n
    \n
    Evaluating the following translations:
    +textbf[{target_language}] translation A: "+textbf[{target_segmentA}]"\n
    +textbf[{target_language}] translation B: "+textbf[{target_segmentB}]"\n
    \n
    Which translation is better? Please output either "A" or "B" only, without any additional 
    explanation.\n
    \n
    Answer:

    (ASSISTANT) +textbf[{choice of verifier}]
    \end{Verbatim}
}
\caption{\textbf{Prompt contexts used in our experiments}. the error analysis evaluator uses three-shot prompting (examples omitted), while both the automatic post-editor and pairwise quality verifier operate in zero-shot mode.}
\label{fig:prompt}
\end{figure*}

\section{Detailed Results} \label{appendix:detail_res}

\subsection{Performance of WMT22}

To facilitate a clearer comparison between MQM and MQM-APE, we present detailed results for each language pair on the WMT22 test set, at both system and segment levels, in Table~\ref{tab:appendix_wmt_res}. This table offers a more comprehensive version of Table~\ref{tab:mainres}.

\begin{table*}[ht]
\small
\centering
\setlength{\tabcolsep}{3pt}
\begin{tabular}{llcccccccc}
\toprule[0.5mm]
\textbf{Model} & \textbf{Prompt} & \textbf{En-De} & \textbf{En-Ru} & \textbf{Zh-En} & \textbf{All.} & \textbf{En-De} & \textbf{En-Ru} & \textbf{Zh-En} & \textbf{Avg.} \\ \midrule
\multirow{2}{*}{\textbf{Llama3-8b-inst}} & \faToggleOff \ MQM & 73.1 & 80.0 & 78.0 & \cellcolor{myblue}77.4 & 54.5 & 48.3 & 45.7 & \cellcolor{myblue}49.5 \\
 & \faToggleOn \ MQM-APE & 80.8 & 85.7 & 82.4 & \cellcolor{myblue}\makecell{83.2 \small\textcolor{red}{(+5.8)}} & 54.4 & 51.0\dag & 47.5\dag & \cellcolor{myblue}\makecell{51.0 \small\textcolor{red}{(+1.5)} \ } \\ \midrule
\multirow{2}{*}{\textbf{Llama3-70b-inst}} & \faToggleOff \ MQM & 80.8 & 83.8 & 82.4 & \cellcolor{myblue}82.5 & 54.3 & 51.7 & 49.4 & \cellcolor{myblue}51.8 \\
 & \faToggleOn \ MQM-APE & 84.6 & 83.8 & 86.8 & \cellcolor{myblue}\makecell{85.0 \small\textcolor{red}{(+2.5)}} & 55.7\dag & 53.3\dag & 50.8\dag & \cellcolor{myblue}\makecell{53.3 \small\textcolor{red}{(+1.5)} \ } \\ \midrule
\multirow{2}{*}{\textbf{Mixtral-8x7b-inst}} & \faToggleOff \ MQM & 80.8 & 90.5 & 84.6 & \cellcolor{myblue}85.8 & 55.2 & 53.5 & 48.8 & \cellcolor{myblue}52.5 \\ 
 & \faToggleOn \ MQM-APE & 84.6 & 87.6 & 84.6 & \cellcolor{myblue}\makecell{85.8 \small(0.0) \ \ } & 55.4 & 53.7 & 50.2\dag & \cellcolor{myblue}\makecell{53.1 \small\textcolor{red}{(+0.6)} \ } \\ \midrule
\multirow{2}{*}{\textbf{Mixtral-8x22b-inst}} & \faToggleOff \ MQM   & 83.3 & 86.7 & 91.2 & \cellcolor{myblue}87.2 & 55.7 & 53.4 & 50.3 & \cellcolor{myblue}53.1 \\
 & \faToggleOn \ MQM-APE & 87.2 & 86.7 & 91.2 & \cellcolor{myblue}\makecell{88.3 \small\textcolor{red}{(+1.1)}} & 56.9\dag & 55.1\dag & 50.6 & \cellcolor{myblue}\makecell{54.2 \small\textcolor{red}{(+1.1)} \ } \\ \midrule
\multirow{2}{*}{\textbf{Qwen1.5-14b-chat}} & \faToggleOff \ MQM   & 85.9 & 82.9 & 83.5 & \cellcolor{myblue}83.9 & 55.6 & 51.0 & 48.2 & \cellcolor{myblue}51.6 \\
 & \faToggleOn \ MQM-APE & 85.9 & 81.9 & 85.7 & \cellcolor{myblue}\makecell{84.3 \small\textcolor{red}{(+0.4)}} & 55.7 & 51.9\dag & 49.8\dag & \cellcolor{myblue}\makecell{52.5 \small\textcolor{red}{(+0.9)} \ } \\ \midrule
\multirow{2}{*}{\textbf{Qwen1.5-72b-chat}} & \faToggleOff \ MQM   & 80.8 & 86.7 & 85.7 & \cellcolor{myblue}84.7 & 56.0 & 54.7 & 50.6 & \cellcolor{myblue}53.8 \\ 
 & \faToggleOn \ MQM-APE & 83.3 & 85.7 & 87.9 & \cellcolor{myblue}\makecell{85.8 \small\textcolor{red}{(+1.1)}} & 56.4 & 55.7 & 51.4\dag & \cellcolor{myblue}\makecell{54.5 \small\textcolor{red}{(+0.7)} \ } \\ \midrule
\multirow{2}{*}{\textbf{Tower-7b-inst}} & \faToggleOff \ MQM   & 73.1 & 90.5 & 79.1 & \cellcolor{myblue}81.8 & 53.6 & 49.6 & 42.1 & \cellcolor{myblue}48.4 \\
 & \faToggleOn \ MQM-APE & 83.3 & 88.6 & 80.2 & \cellcolor{myblue}\makecell{84.3 \small\textcolor{red}{(+2.5)}} & 53.5 & 50.9\dag & 44.4\dag & \cellcolor{myblue}\makecell{49.6 \small\textcolor{red}{(+1.2)} \ } \\ \midrule
\multirow{2}{*}{\textbf{Tower-13b-inst}} & \faToggleOff \ MQM   & 75.6 & 95.2 & 76.9 & \cellcolor{myblue}83.6 & 55.8 & 55.7 & 44.9 & \cellcolor{myblue}52.1 \\
 & \faToggleOn \ MQM-APE & 75.6 & 96.2 & 80.2 & \cellcolor{myblue}\makecell{85.0 \small\textcolor{red}{(+1.4)}} & 56.3 & 55.7 & 45.0 &\cellcolor{myblue} \makecell{52.3 \small\textcolor{red}{(+0.2)}} \\ \bottomrule[0.5mm]
\end{tabular}
\caption{\textbf{The WMT22 performance of GEMBA-MQM ("MQM") vs. MQM-APE} using pairwise accuracy (\%) at the system level and pairwise accuracy with tie calibration (\%) at the segment level. All results are compared with human-annotated MQM scores. "\dag" denotes cases where MQM-APE is significantly better than GEMBA-MQM.}
\label{tab:appendix_wmt_res}
\end{table*}

\subsection{Performance of IndicMT}

To offer a more detailed comparison between MQM and MQM-APE, we present results for each language on the IndicMT test set, in Table~\ref{tab:appendix_indicmt_res}. This table offers a more comprehensive version of Table~\ref{tab:indicmt_res}.

\begin{table*}[ht]
\centering
\small
\begin{tabular}{llccccc}
\toprule[0.5mm]
\textbf{Model} & \textbf{Prompt} & \textbf{Assamese} & \textbf{Maithili} & \textbf{Kannada} & \textbf{Punjabi} & \textbf{Avg.} \\\midrule
\multirow{2}{*}{\textbf{Llama3-8b-inst}} & \faToggleOff \ MQM & 37.8 & 18.1 & 41.3 & 39.2 & \cellcolor{myblue}34.1 \\
& \faToggleOn \ MQM-APE & 43.5 & 41.7 & 41.7 & 38.9 & \cellcolor{myblue}41.5 \textcolor{red}{(+7.4)} \\\midrule
\multirow{2}{*}{\textbf{Llama3-70b-inst}} & \faToggleOff \ MQM & 40.1 & 32.1 & 41.3 & 41.3 & \cellcolor{myblue}38.7 \\
& \faToggleOn \ MQM-APE & 46.4 & 46.1 & 42.6 & 41.4 & \cellcolor{myblue}44.1 \textcolor{red}{(+5.4)} \\\midrule
\multirow{2}{*}{\textbf{Mixtral-8x7b-inst}} & \faToggleOff \ MQM & 35.1 & 22.2 & 43.0 & 42.5 & \cellcolor{myblue}35.7 \\
& \faToggleOn \ MQM-APE & 42.0 & 35.6 & 43.2 & 41.4 & \cellcolor{myblue}40.5 \textcolor{red}{(+4.8)} \\\midrule
\multirow{2}{*}{\textbf{Mixtral-8x22b-inst}} & \faToggleOff \ MQM & 41.8 & 44.8 & 51.5 & 37.8 & \cellcolor{myblue}44.0 \\
& \faToggleOn \ MQM-APE & 46.2 & 46.2 & 50.2 & 39.1 & \cellcolor{myblue}45.4 \textcolor{red}{(+1.4)} \\\midrule
\multirow{2}{*}{\textbf{Qwen1.5-14b-chat}} & \faToggleOff \ MQM & 21.5 & 8.6 & 24.8 & 37.0 & \cellcolor{myblue}23.0 \\
& \faToggleOn \ MQM-APE & 39.1 & 20.7 & 50.1 & 39.7 & \cellcolor{myblue}37.4 \textcolor{red}{(+14.4)} \\\midrule
\multirow{2}{*}{\textbf{Qwen1.5-72b-chat}} & \faToggleOff \ MQM & 46.0 & 37.7 & 48.9 & 43.1 & \cellcolor{myblue}43.9 \\
& \faToggleOn \ MQM-APE & 45.4 & 46.9 & 43.7 & 43.6 & \cellcolor{myblue}44.9 \textcolor{red}{(+1.0)} \\\midrule
\multirow{2}{*}{\textbf{Tower-7b-inst}} & \faToggleOff \ MQM & 23.7 & 13.3 & 28.6 & 38.6 & \cellcolor{myblue}26.0 \\
& \faToggleOn \ MQM-APE & 33.3 & 23.2 & 38.6 & 37.6 & \cellcolor{myblue}33.2 \textcolor{red}{(+7.2)} \\\midrule
\multirow{2}{*}{\textbf{Tower-13b-inst}} & \faToggleOff \ MQM & 36.7 & 17.8 & 43.5 & 38.0 & \cellcolor{myblue}34.0 \\
& \faToggleOn \ MQM-APE & 40.0 & 16.9 & 42.6 & 38.3 & \cellcolor{myblue}34.5 \textcolor{red}{(+0.5)} \\
\bottomrule[0.5mm]
\end{tabular}
\caption{\textbf{Segment-level performance of GEMBA-MQM("MQM") vs. MQM-APE} on IndicMT dataset. All results are compared with human-annotated MQM scores.}
\label{tab:appendix_indicmt_res}
\end{table*}

\subsection{Performance of Error Quality}

To provide a more detailed comparison of predicted error quality between MQM and MQM-APE, we present performance measured by SP and MP for each language pair in Table~\ref{tab:appendix_error_res}. This table expands upon the error quality results shown in Table~\ref{tab:mainres}.

\begin{table*}[ht]
\centering
\small
\begin{tabular}{llcccccccc}
\toprule[0.5mm]
 \multirow{2}{*}{\textbf{Model}} & \multirow{2}{*}{\textbf{Prompt}} & \multicolumn{2}{c}{\textbf{En-De}} & \multicolumn{2}{c}{\textbf{En-Ru}} & \multicolumn{2}{c}{\textbf{Zh-En}} & \multicolumn{2}{c}{\textbf{Avg.}} \\
 \cmidrule(lr){3-4} \cmidrule(lr){5-6} \cmidrule(lr){7-8} \cmidrule(lr){9-10}
 & & \textbf{SP} & \textbf{MP} & \textbf{SP} & \textbf{MP} & \textbf{SP} & \textbf{MP} & \textbf{SP} & \textbf{MP} \\\midrule
\multirow{2}{*}{\textbf{Llama3-8b-inst}} & \faToggleOff \ MQM & 7.61 & 3.28 & 9.72 & 4.93 & 9.75 & 6.44 & \cellcolor{myblue}9.03 & \cellcolor{myblue} 4.88 \\
& \faToggleOn \ MQM-APE & 8.87\dag & 4.07\dag & 11.37\dag & 5.72\dag & 11.02\dag & 7.06\dag & \cellcolor{myblue}10.42 & \cellcolor{myblue}5.62 \\\midrule
\multirow{2}{*}{\textbf{Llama3-70b-inst}} & \faToggleOff \ MQM & 14.46 & 6.95 & 16.67 & 10.18 & 17.57 & 16.29 & \cellcolor{myblue}16.23 & \cellcolor{myblue}11.14 \\
& \faToggleOn \ MQM-APE & 15.35\dag & 7.43\dag & 18.29\dag & 10.99\dag & 17.93\dag & 16.64\dag & \cellcolor{myblue}17.19 & \cellcolor{myblue}11.69 \\\midrule
\multirow{2}{*}{\textbf{Mixtral-8x7b-inst}} & \faToggleOff \ MQM & 9.11 & 3.25 & 10.66 & 2.22 & 11.69 & 9.54 & \cellcolor{myblue}10.49 & \cellcolor{myblue}5.00 \\
& \faToggleOn \ MQM-APE & 9.41\dag & 3.36\dag & 10.87\dag & 2.25\dag & 11.94\dag & 9.72\dag & \cellcolor{myblue}10.74 & \cellcolor{myblue}5.11 \\\midrule
\multirow{2}{*}{\textbf{Mixtral-8x22b-inst}} & \faToggleOff \ MQM & 8.66 & 5.34 & 9.73 & 4.42 & 11.37 & 10.47 & \cellcolor{myblue}9.92 & \cellcolor{myblue}6.74 \\
& \faToggleOn \ MQM-APE & 9.22\dag & 5.48\dag & 10.25\dag & 4.62\dag & 11.50\dag & 10.58\dag & \cellcolor{myblue}10.32 & \cellcolor{myblue}6.89 \\\midrule
\multirow{2}{*}{\textbf{Qwen1.5-14b-chat}} & \faToggleOff \ MQM & 8.03 & 2.47 & 8.36 & 2.68 & 13.12 & 8.56 & \cellcolor{myblue}9.84 & \cellcolor{myblue}4.57 \\
& \faToggleOn \ MQM-APE & 8.44\dag & 2.71\dag & 8.67\dag & 2.67 & 13.82\dag & 8.93\dag & \cellcolor{myblue}10.31 & \cellcolor{myblue}4.77 \\\midrule
\multirow{2}{*}{\textbf{Qwen1.5-72b-chat}} & \faToggleOff \ MQM & 7.90 & 2.32 & 9.99 & 3.05 & 9.76 & 4.59 & \cellcolor{myblue}9.22 & \cellcolor{myblue}3.32 \\
& \faToggleOn \ MQM-APE & 8.95\dag & 2.56\dag & 11.25\dag & 3.42\dag & 10.80\dag & 5.15\dag & \cellcolor{myblue}10.33 & \cellcolor{myblue}3.71 \\\midrule
\multirow{2}{*}{\textbf{Tower-7b-inst}} & \faToggleOff \ MQM & 11.31 & 1.54 & 21.44 & 2.91 & 18.83 & 7.96 & \cellcolor{myblue}17.19 & \cellcolor{myblue}4.14 \\
& \faToggleOn \ MQM-APE & 11.49\dag & 1.60\dag & 21.98\dag & 2.94 & 19.12\dag & 8.37\dag & \cellcolor{myblue}17.53 & \cellcolor{myblue}4.30 \\\midrule
\multirow{2}{*}{\textbf{Tower-13b-inst}} & \faToggleOff \ MQM & 15.80 & 9.87 & 27.73 & 24.09 & 19.96 & 14.38 & \cellcolor{myblue}21.16 & \cellcolor{myblue}16.11 \\
& \faToggleOn \ MQM-APE & 15.96\dag & 10.04\dag & 28.07\dag & 23.72 & 20.26\dag & 13.59 & \cellcolor{myblue}21.43 & \cellcolor{myblue}15.78 \\
\bottomrule[0.5mm]
\end{tabular}
\caption{\textbf{Comparison between GEMBA-MQM ("MQM") and MQM-APE on error span quality evaluation} for the WMT22 test set, measured using span precision (SP) and major error span precision (MP). "\dag" indicates cases where the error span quality from MQM-APE significantly surpasses that of GEMBA-MQM. }
\label{tab:appendix_error_res}
\end{table*}

\section{Detailed Analysis of Inference Cost} \label{appendix:infer_cost}

Table~\ref{tab:appendix_inferencecost} shows the inference costs for different LLMs using MQM-APE. Although the input sizes across the various LLMs are similar, there is a significant variance in the number of tokens generated during inference. Specifically, the Qwen1.5 series models generate fewer tokens compared to others, which may result in lower inference costs. In contrast, the Mixtral model produces significantly more tokens, likely due to this model's stricker error detection to identify more errors during the initial Error Analysis phase.

\begin{table*}[ht]
\centering
\small
\begin{tabular}{lcccccccc}
\toprule[0.5mm]
\multirow{2}{*}{\textbf{Model}} & \multicolumn{2}{c}{\textbf{Error Analysis}} & \multicolumn{2}{c}{\textbf{Error-based APE}} & \multicolumn{2}{c}{\textbf{Quality Verifier}} & \multicolumn{2}{c}{\textbf{Total}} \\
\cmidrule(lr){2-3} \cmidrule(lr){4-5} \cmidrule(lr){6-7} \cmidrule(lr){8-9}
& \textbf{Input} & \textbf{Generated} & \textbf{Input} & \textbf{Generated} & \textbf{Input} & \textbf{Generated} & \textbf{Input} & \textbf{Generated} \\\midrule
\textbf{Llama3-8b-inst} & 1168.9 & 52.2 & 207.7 & 61.7 & 248.7 & 1.5 & 1378.9 & 113.4 \\
\textbf{Llama3-70b-inst} & 1168.9 & 34.9 & 136.1 & 43.7 & 166.7 & 1.0 & 1305.0 & 79.6 \\\midrule
\textbf{Mixtral-8x7b-inst} & 1390.0 & 71.6 & 485.7 & 212.2 & 619.8 & 55.5 & 1875.7 & 339.3 \\
\textbf{Mixtral-8x22b-inst} & 1368.0 & 63.6 & 353.7 & 136.9 & 449.4 & 11.0 & 1721.7 & 211.5 \\\midrule
\textbf{Qwen1.5-14b-chat} & 1168.4 & 20.0 & 110.2 & 36.0 & 135.9 & 0.7 & 1278.6 & 56.7 \\
\textbf{Qwen1.5-72b-chat} & 1168.4 & 38.6 & 188.5 & 53.4 & 222.8 & 1.3 & 1357.0 & 93.3 \\\midrule
\textbf{Tower-13b-inst} & 1466.3 & 32.5 & 312.4 & 98.9 & 396.2 & 18.6 & 1778.7 & 150.0 \\
\textbf{Tower-7b-inst} & 1466.3 & 60.1 & 332.9 & 127.6 & 403.4 & 13.9 & 1799.2 & 201.6 \\\midrule
\cellcolor{myblue} \textbf{Average Cost} & \cellcolor{myblue} 1295.7 & \cellcolor{myblue} 46.7 & \cellcolor{myblue} 265.9 & \cellcolor{myblue} 96.3 & \cellcolor{myblue} 349.5 & \cellcolor{myblue} 12.7 & \cellcolor{myblue} 1736.2 & \cellcolor{myblue} 144.4 \\
\bottomrule[0.5mm]
\end{tabular}
\caption{\textbf{Analysis of inference cost} averaged for each segment across different LLMs for each module, presenting input and generated tokens separately.}
\label{tab:appendix_inferencecost}
\end{table*}

\section{Detailed Analysis of Severity and Category Distribution of Errors} \label{appendix:error_analysis}

\subsection{Error Severity Distribution}

Table~\ref{tab:appendix_error_severity} shows the number of errors identified. The results are also presented in Figure~\ref{fig:errors_compare}. Apart from the main findings that MQM-APE preserves a similar distribution compared with original evaluator, we also observe that Mixtral outputs more errors than other models. An interesting insight for Tower-7b-inst is that most of the generated errors are considered as unimpactful and discarded, showing that Tower-7b-inst identified errors that are not that reliable from the original evaluator.

\subsection{Error Category Distribution}

\begin{figure*}[ht]
\includegraphics[scale=0.43]{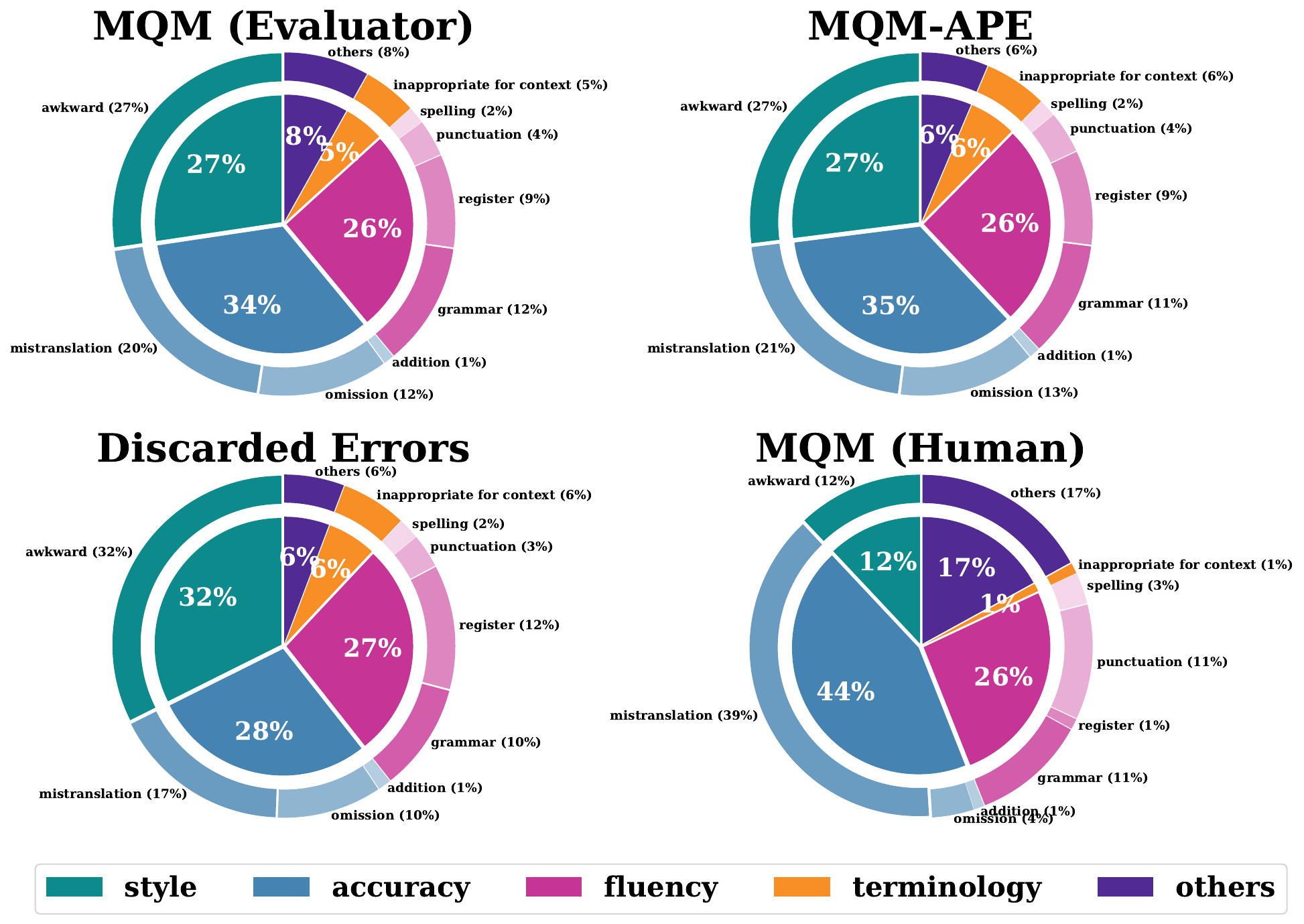}
\centering
\caption{\textbf{Distribution of error categories} generated from GEMBA-MQM ("MQM") evaluator, MQM-APE, discarded errors, and human-annotated MQM, respectively.}
\label{fig:appendix_category_pie}
\end{figure*}

Figure~\ref{fig:appendix_category_pie} shows the distribution of error categories from the MQM evaluator, MQM-APE, discarded errors, and human annotations. Apart from the main finding that MQM-APE preserves the distribution of error categories, there is a notable misalignment between the LLM evaluator and human annotations. Specifically, 27\% of errors identified by the LLM evaluator fall into the "style/awkward" category, whereas human MQM annotations focus more on "accuracy/mistranslation". We recommend that future evaluators emphasize mistranslation detection and exercise caution with style-related errors.

Table~\ref{tab:appendix_error_category} and Table~\ref{tab:appendix_error_category_filter} compare the top 3 generated or discarded errors for each LLM. One key finding is that discarded errors are predominantly categorized as 'style/awkward,' suggesting that many of these errors may be less reliable in GEMBA-MQM. Another observation is that many terminology errors are discarded by MQM-APE for the Mixtral-8x22b-inst models, suggesting that MQM-APE reduces the bias in generating style and terminology errors in the original error distribution.

Table~\ref{tab:appendix_error_category_top7} presents the top 7 error categories generated by evaluators and human annotations. Notably, certain human-annotated errors, such as fluency/spelling and inconsistency, are rarely identified by LLM-based evaluators. Future work should focus on adjusting the error distribution to better capture these long-tailed errors.

\begin{table*}[ht]
\centering
\small
\begin{tabular}{lcccccccc}
\toprule[0.5mm]
\multirow{2}{*}{\textbf{Model}} & \multicolumn{2}{c}{\textbf{Critical}} & \multicolumn{2}{c}{\textbf{Major}} & \multicolumn{2}{c}{\textbf{Minor}} & \multicolumn{2}{c}{\textbf{Total}} \\
\cmidrule(lr){2-3} \cmidrule(lr){4-5} \cmidrule(lr){6-7} \cmidrule(lr){8-9}
& \textbf{Origin} & \textbf{Remain} & \textbf{Origin} & \textbf{Remain} & \textbf{Origin} & \textbf{Remain} & \textbf{Origin} & \textbf{Remain} \\\midrule
\textbf{Llama3-8b-inst}     & 0.20 & 0.14 & 0.52 & 0.26 & 0.76 & 0.36 & \cellcolor{myblue} 1.49 & \cellcolor{myblue} 0.75 \\
\textbf{Llama3-70b-inst}    & 0.06 & 0.05 & 0.46 & 0.36 & 0.50 & 0.34 & \cellcolor{myblue} 1.02 & \cellcolor{myblue} 0.75 \\ \midrule
\textbf{Mixtral-8x7b-inst}  & 0.15 & 0.11 & 0.25 & 0.17 & 2.43 & 1.59 & \cellcolor{myblue} 2.83 & \cellcolor{myblue} 1.87 \\
\textbf{Mixtral-8x22b-inst} & 0.09 & 0.07 & 0.40 & 0.34 & 1.72 & 1.27 & \cellcolor{myblue} 2.20 & \cellcolor{myblue} 1.68 \\ \midrule
\textbf{Qwen1.5-14b-chat}    & 0.15 & 0.13 & 0.18 & 0.14 & 0.33 & 0.23 & \cellcolor{myblue} 0.67 & \cellcolor{myblue} 0.50 \\
\textbf{Qwen1.5-72b-chat}    & 0.14 & 0.13 & 0.35 & 0.26 & 0.81 & 0.52 & \cellcolor{myblue} 1.30 & \cellcolor{myblue} 0.91 \\ \midrule
\textbf{Tower-7b-inst}      & 0.85 & 0.35 & 0.81 & 0.33 & 0.34 & 0.15 & \cellcolor{myblue} 2.00 & \cellcolor{myblue} 0.83 \\
\textbf{Tower-13b-inst}     & 0.13 & 0.07 & 0.09 & 0.05 & 1.53 & 0.76 & \cellcolor{myblue} 1.75 & \cellcolor{myblue} 0.88 \\ \midrule
\cellcolor{mygold} \textbf{MQM (Human)} & \cellcolor{mygold} 0.00 & \cellcolor{mygold} - & \cellcolor{mygold} 0.26 & \cellcolor{mygold} - & \cellcolor{mygold} 0.41 & \cellcolor{mygold} - & \cellcolor{mygold} 0.67 & \cellcolor{mygold} - \\
\bottomrule[0.5mm]
\end{tabular}
\caption{\textbf{Comparison of the average number of errors per segment} before and after applying MQM-APE, categorized by severity as "Critical", "Major", and "Minor".}
\label{tab:appendix_error_severity}
\end{table*}

\begin{table*}[ht]
\centering
\scriptsize
\setlength{\tabcolsep}{2pt}
\begin{tabular}{lcccc}
\toprule[0.5mm]
\textbf{Model} & \textbf{Top1 Category (\%)} & \textbf{Top2 Category (\%)} & \textbf{Top3 Category (\%)} & \textbf{Others (\%)} \\\midrule
\textbf{Llama3-8b-inst}   & accuracy/mistranslation (43\%) & fluency/grammar (40\%)   & style/awkward (6\%)                           & 10\%           \\ \midrule
\textbf{Llama3-70b-inst}  & accuracy/mistranslation (46\%) & fluency/register (21\%)  & style/awkward (10\%)                          & 23\%           \\ \midrule
\textbf{Mixtral-8x7b-inst}  & style/awkward (24\%)          & fluency/register (24\%)  & \makecell{terminology/inappropriate \\ for context (15\%)} & 37\%           \\ \midrule
\textbf{Mixtral-8x22b-inst} & style/awkward (29\%)          & fluency/register (18\%)  & accuracy/mistranslation (15\%)               & 38\%           \\ \midrule
\textbf{Qwen1.5-14b-chat}   & style/awkward (31\%)          & accuracy/omission (24\%) & accuracy/mistranslation (20\%)               & 25\%           \\ \midrule
\textbf{Qwen1.5-72b-chat}   & style/awkward (37\%)          & accuracy/omission (26\%) & accuracy/mistranslation (12\%)               & 25\%           \\ \midrule
\textbf{Tower-7b-inst}     & style/awkward (30\%)          & accuracy/omission (24\%) & accuracy/mistranslation (13\%)               & 33\%           \\ \midrule
\textbf{Tower-13b-inst}    & style/awkward (52\%)          & fluency/grammar (21\%)   & style/inconsistent use (15\%)                & 12\%           \\ \midrule
\cellcolor{mygold} \textbf{MQM (Human)} & \cellcolor{mygold} accuracy/mistranslation(39\%) & \cellcolor{mygold} style/awkward(12\%) & \cellcolor{mygold} fluency/grammar(9\%) & \cellcolor{mygold} 40\% \\
\bottomrule[0.5mm]
\end{tabular}
\caption{\textbf{Top 3 error categories generated by MQM-prompted evaluators} from different LLMs, compared to human-annotated MQM.}
\label{tab:appendix_error_category}
\end{table*}

\begin{table*}[ht]
\centering
\scriptsize
\setlength{\tabcolsep}{0.3pt}
\begin{tabular}{lcccc}
\toprule[0.5mm]
\textbf{Model} & \textbf{Top1 Category (\%)} & \textbf{Top2 Category (\%)} & \textbf{Top3 Category (\%)} & \textbf{Others (\%)} \\\midrule
\textbf{Llama3-8b-inst}  & fluency/grammar (47\%) & accuracy/mistranslation (40\%) & style/awkward (6\%) & 7\% \\ \midrule
\textbf{Llama3-70b-inst} & accuracy/mistranslation (41\%) & fluency/register (35\%) & style/awkward (7\%) & 17\% \\ \midrule
\textbf{Mixtral-8x7b-inst} & fluency/register (25\%) & style/awkward (25\%) & \makecell{terminology/inappropriate \\ for context (17\%)} & 33\% \\ \midrule
\textbf{Mixtral-8x22b-inst} & \makecell{terminology/inappropriate \\ for context (27\%)} & style/awkward (27\%) & fluency/register (22\%) & 23\% \\ \midrule
\textbf{Qwen15-14b-chat} & style/awkward (45\%) & accuracy/omission (17\%) & accuracy/mistranslation (14\%) & 24\% \\ \midrule
\textbf{Qwen15-72b-chat} & style/awkward (50\%) & accuracy/omission (23\%) & fluency/register (8\%) & 19\% \\ \midrule
\textbf{Tower-7b-inst} & style/awkward (33\%) & accuracy/omission (26\%) & accuracy/mistranslation (17\%) & 24\% \\ \midrule
\textbf{Tower-13b-inst} & style/awkward (66\%) & fluency/grammar (16\%) & style/inconsistent use (7\%) & 11\% \\
\bottomrule[0.5mm]
\end{tabular}
\caption{\textbf{Top 3 error categories discarded by MQM-APE} from different LLMs.}
\label{tab:appendix_error_category_filter}
\end{table*}

\begin{table*}[ht]
\centering
\scriptsize
\begin{tabular}{lcccc}
\toprule[0.5mm]
\textbf{Top.} & \textbf{MQM (Evaluator) (\%)} & \textbf{MQM-APE (\%)} & \textbf{Discarded Errors (\%)} & \textbf{MQM (Human) (\%)} \\\midrule
1 & style/awkward (27\%) & style/awkward (27\%) & style/awkward (32\%) & accuracy/mistranslation (44\%) \\\midrule
2 & accuracy/mistranslation (20\%) & accuracy/mistranslation (21\%) & accuracy/mistranslation (17\%) & style/awkward (12\%) \\\midrule
3 & accuracy/omission (12\%) & accuracy/omission (13\%) & fluency/register (12\%) & fluency/grammar (11\%) \\\midrule
4 & fluency/grammar (12\%) & fluency/grammar (11\%) & fluency/grammar (10\%) & fluency/punctuation (11\%) \\\midrule
5 & fluency/register (9\%) & fluency/register (9\%) & accuracy/omission (10\%) & accuracy/omission (4\$) \\\midrule
6 & \makecell{terminology/inappropriate \\ for context (5\%)} & \makecell{terminology/inappropriate \\ for context (5\%)} & \makecell{terminology/inappropriate \\ for context (6\%)} & fluency/spelling (3\%) \\\midrule
7 & fluency/punctuation (4\%) & fluency/punctuation (4\%) & fluency/punctuation (3\%) & fluency/inconsistency (3\%) \\
\bottomrule[0.5mm]
\end{tabular}
\caption{\textbf{Top 7 error categories} generated from GEMBA-MQM ("MQM Evaluator"), MQM-APE, discarded errors, and human-annotated MQM, respectively.}
\label{tab:appendix_error_category_top7}
\end{table*}

\section{Positional Bias and Invalid Answers} \label{appendix:bias_and_invalid}

Previous studies have shown that pairwise evaluation may be influenced by positional bias \citep{shi2024judging}. To reduce this effect in our approach, we run the pairwise quality verifier twice, swapping the translations each time. We apply a simple 'smoothing' technique when calculating the final scores, assigning half-weight to contrastive results.

Since our approach involves three types of LLM-based inferences, there is still a chance that the LLMs' responses may not be formatted as expected. To address this issue, we follow \citet{kocmi-federmann-2023-large} and regenerate the answer by slightly increasing the temperature. Fortunately, we encountered no significant instances of invalid responses across the LLMs tested. As instruction-tuning capabilities continue to improve in future models, we expect this issue to become even less relevant.

\end{document}